\documentclass[preprint,12pt]{elsarticle}

\usepackage{graphicx}
\usepackage{amssymb}

\usepackage{lineno}

\usepackage[utf8]{inputenc}
\usepackage{lmodern}
\usepackage[T1]{fontenc}
\usepackage{booktabs}
\usepackage{lscape}
\usepackage{amsmath}	
\usepackage{cases}
\usepackage{caption}
\usepackage{comment}

\usepackage{makecell}
\usepackage{multirow}
\usepackage{amsfonts}
\usepackage{float} 
\usepackage{diagbox}
\usepackage{fullpage}
\usepackage{url}
\usepackage{rotating}
\usepackage{eurosym}
\usepackage{wrapfig}
\usepackage[final]{pdfpages} 
\usepackage{epstopdf} 
\usepackage[a4paper]{geometry}
\usepackage[nottoc]{tocbibind}
\usepackage{placeins}
\usepackage{listings}
\usepackage{color, colortbl}
\usepackage{hyperref}
\usepackage{xcolor}
\usepackage{caption, subcaption}
\usepackage[normalem]{ulem}  

\usepackage{sidecap}
\hypersetup{
colorlinks,
citecolor=black,
filecolor=black,
linkcolor=black,
urlcolor=black
}

\newcommand{\firstrevision}{\textcolor{black}}
\newcommand{\secondrevision}{\textcolor{black}}
\newcommand{\thirdrevision}{\textcolor{black}}
\newcommand{\fourthrevision}{\textcolor{black}}

\newcommand{\bH}{\mathbf{H}}
\newcommand{\bR}{\mathbf{R}}

\newcommand{\bQ}{\mathbf{Q}}
\newcommand{\bY}{\mathbf{Y}}

\newcommand{\bV}{\mathbf{V}}

\newcommand{\bU}{\mathbf{U}}

\newcommand{\bS}{\mathbf{S}}

\newcommand{\bP}{\mathbf{P}}
\newcommand{\bPsi}{\boldsymbol{\Psi}}

\journal{Building and Environment}

\begin{document}

\begin{frontmatter}


    \title{Rooftop Wind Field Reconstruction Using Sparse Sensors: From Deterministic to Generative Learning Methods}

    \author{Yihang Zhou\texorpdfstring{$^{1,\dagger}$}{(1,†)}, Chao Lin\texorpdfstring{$^{2,\dagger}$}{(2,†)}, Hideki Kikumoto\texorpdfstring{$^{2}$}{(2)}, Ryozo Ooka\texorpdfstring{$^{2}$}{(2)}, Sibo Cheng \texorpdfstring{$^{3,*}$}{(3,*)} \\ \texorpdfstring{$^{\dagger}$}{†} co-first author, \texorpdfstring{$^{*}$}{*} corresponding: sibo.cheng@enpc.fr \\
        \small \texorpdfstring{$^{1}$}{(1)}Department of Computing, Imperial College London, London, UK \\
        \small \texorpdfstring{$^{2}$}{(2)}Institute of Industrial Science, The University of Tokyo, Tokyo, Japan \\
        \small \texorpdfstring{$^{3}$}{(3)}CEREA, ENPC, EDF R\&D, Institut Polytechnique de Paris, \^Ile-de-France, France \\
    }


    \section*{Highlights}
    \vspace{-0.5em}
    \begin{itemize}
        \item Comparing deterministic and generative learning methods against Kriging interpolation.
        \item Evaluating methods and strategies under real-world constraints using experimental data.
        \item Investigating a sensor position optimization method and its effectiveness.
        \item Providing method selection advice under various application scenarios.
    \end{itemize}
    \vspace{1em}

    \begin{abstract}

        Real-time rooftop wind-speed distribution is essential for safe operation of drones or urban air mobilities, wind control systems, and rooftop utilization. However, rooftop flows exhibit strong nonlinearity, separation, and cross-direction variability, which challenge flow field reconstruction from sparse sensors. This study develops a learning-from-observation framework using wind-tunnel experiment data obtained by Particle Image Velocimetry (PIV) and benchmarks Kriging interpolation against three deep learning models—UNet, Vision Transformer Autoencoder (ViTAE), and Conditional Wasserstein GAN (CWGAN). We evaluate two training strategies—single  wind-direction training (SDT) and mixed wind-direction training (MDT)—across sensor densities of 5–30, test robustness under ±1-grid sensor position perturbations, and optimize sensor placement via Proper Orthogonal Decomposition with QR decomposition. Results indicate the potential of deep learning methods to reconstruct rooftop wind fields from sparse sensor data. Compared with traditional Kriging interpolation, the deep learning models achieved improvements of up to 32.7\% in SSIM, 24.2\% in FAC2, and 27.8\% in NMSE. In addition, training with mixed wind directions was found to be essential for achieving better model performance, resulting in further improvements of up to 173.7\% in SSIM, 16.7\% in FAC2, and 98.3\% in MG compared with single-direction training. Furthermore, the comparison among methods indicates that sensor configuration, optimization, and training strategy should be jointly considered to ensure reliable deployment. For example, QR-based optimization enhances robustness by up to 27.8\% under sensor perturbations, although with metric-dependent performance trade-offs. By training on experimental rather than simulated data, the proposed framework also offers practical guidance for method selection and sensor placement across different scenarios. The source code for this project is publicly available as open source on Github ( \url{https://github.com/Yng314/windreconstruction}).

    \end{abstract}

    \begin{keyword}
        Wind field reconstruction \sep Sparse sensors \sep Deep learning \sep PIV measurement \sep Sensor optimization
    \end{keyword}

\end{frontmatter}


\begin{table}[h!]
    \raggedright
    \begin{tabular}{cl}
        \textbf{Main Notations}      &                                                     \\
        \\
        $u$, $v$                     & Velocity components in $x$, $y$ directions          \\
        $w_s$                        & Normalized horizontal wind speed                    \\
        $U_H$                        & Reference velocity at model height $H$              \\
        $WS$, $ws_{SD}$              & Time-averaged wind speed and its standard deviation \\
        $(x, y, z)$                  & Spatial coordinates                                 \\
        $H$                          & Model height                                        \\
        $\Delta t$                   & Time step                                           \\
        $\bH$                        & Observation operator matrix                         \\
        $\bPsi$, $\bPsi_r$           & POD basis matrix and reduced basis with $r$ modes   \\
        $\bY$                        & Wind field data matrix                              \\
        $\bU$, $\bS$, $\bV$          & SVD decomposition matrices                          \\
        $\bQ$, $\bR$                 & QR decomposition matrices                           \\
        $\bP$                        & Permutation matrix                                  \\
        $\mathbf{p}$                 & Permutation vector                                  \\
        $p_1$                        & Most informative sensor location                    \\
        $\mathcal{D}$                & Dataset                                             \\
        $\mathcal{D}_{\theta}^{(k)}$ & $k$-th realization for direction $\theta$           \\
        $N$                          & Number of temporal snapshots                        \\
        $r$                          & Number of retained POD modes                        \\
        $\theta$                     & Wind direction angle                                \\
        $n_\theta$                   & Number of realizations for direction $\theta$       \\
        $O_i$, $P_i$                 & Observed and predicted values at location $i$       \\
        $W$                          & Error tolerance threshold                           \\
    \end{tabular}
\end{table}

\clearpage

\section{Introduction}

Rooftop spaces serve multiple functions in modern urban environments, including building-integrated photovoltaic \cite{maurer2023comparing}, rooftop gardens \cite{fleck2022urban} and HVAC equipment placement \cite{doddipatla2021wind}. With the rapid development of drone and urban air mobility in recent years, vertical takeoff and landing (VTOL) operations have emerged as an increasingly important consideration for rooftop space utilization \cite{watkins2020ten, castagno2021map}.

However, the spatiotemporal variations of rooftop wind environments are highly complex due to building geometry effects and aerodynamic interactions characterized by separation flows and conical flows under different wind directions \cite{pu2023research, carpentieri2015influence}. These uneven flow patterns directly pose danger to the safe utilization of rooftop spaces \cite{Tabrizi2014Performance} and therefore accurate real-time flow field information is expected for operational decision-making, particularly for drone operations and wind control system adjustments \cite{gianfelice2022real, Krawczyk2025Urban}.

Traditional approaches for wind environment characterization include field measurements and Computational Fluid Dynamics (CFD) simulations \cite{Yazid2014A, gnatowska2017cfd}. Field measurement methods provide accurate local information but are limited by the number of observation points and spatial coverage constraints \cite{aitken2014large}. Although CFD methods provide full-field flow information, their practical application is constrained by high computational cost, sensitivity to boundary conditions, lack of real-time capability, and challenges in accurately capturing realistic turbulence characteristics \cite{Shao2023PIGNN-CFD:,Tominaga2024CFD, Hooff2017On}. These limitations have motivated the development of wind field reconstruction approaches that combine sparse sensor measurements with interpolation or machine learning techniques to obtain real-time full-field wind information \cite{Shao2023PIGNN-CFD:, Lin2020Kriging, Lin2021Nonstationary, Gao2024Urban, cheng2025machine}.

Current wind field reconstruction methods in urban environments include conventional approaches such as Proper Orthogonal Decomposition with Linear Stochastic Estimation (POD-LSE) \cite{hu2022estimation, hu2023estimation,riva2024multi,liu2024application} and deep learning methods such as generative adversarial networks \cite{zhang2022towards, hu2024fast, gao2024sigan, hou2024machine}. However, existing approaches face significant limitations from both data and methodological perspectives. One of the critical limitations is the widespread reliance on CFD simulation data for model training rather than experimental observations \cite{Kang2021Application, Ti2020Wake, DeOliveira2022Coupling}. \fourthrevision{Validated CFD has demonstrated strong agreement with experimental measurements for rooftop wind flows. However, even well-validated results may contain systematic biases associated with turbulence closure models and discretization schemes \cite{eidi2022data}, which can alter the statistical characteristics of the training data. In contrast, wind tunnel measurements inherently capture real-world turbulence variability and measurement noise. Since the target application involves real-world deployment where measurement noise is unavoidable, training on experimental data provides improved robustness to such disturbances compared to training on CFD-generated data \cite{kohler2019toward}.} On the other hand, sensor measurements contain real noise and therefore flow fields exhibit natural fluctuations absent in computational simulations \cite{chowdhury2024state,cammi2024data}. Additionally, some studies employ direction-specific training strategies where models are trained and tested on identical wind directions \cite{Qin2018Wind, Hu2024Effect, zhang2022towards}, limiting cross-directional generalization capabilities across diverse flow conditions. \secondrevision{Finally, these simulation-based studies also overlook several real world constraints. For example, they usually do not consider sensor position offsets or sensor failures caused by various reasons. Moreover, they generally assume to predict based on sufficient data, however, in practice, data acquisition sometimes can be challenging, and the performance of models under limited-data conditions remains unevaluated.}

From a methodological perspective, traditional dimensionality reduction methods require extensive training data and struggle with nonlinear flow features \cite{Lawson2010Understanding,Li2013Model,Ferrero2018Global}, while conventional interpolation techniques such as Kriging interpolation assume spatial stationarity that is often violated in complex rooftop environments with flow separation and recirculation zones \cite{Miao2024Interpolation,Risser2016Review:}. To address these limitations, deep learning methods have been introduced in this field. Deep learning approaches have demonstrated strong performance in image reconstruction tasks \cite{koetzier2023deep,An2024A} and offer advantages for handling nonlinear flow field features and long-range spatial dependencies \cite{Miyanawala2018A,cheng2025machine}. Different neural network architectures, including convolutional neural networks \cite{tschannen2018recent}, generative adversarial networks \cite{goodfellow2014generativeadversarialnetworks}, and Vision Transformers \cite{vaswani2023attentionneed}, provide distinct capabilities for feature extraction and reconstruction. Compared to traditional interpolation methods, deep learning architectures can learn complex mappings between sparse and unstructured sensor measurements \cite{cheng2025machine,cheng2024efficient}, and full flow fields without explicit assumptions about spatial correlation structures \cite{Wang2020An, Szczotka2019Learning, Makarov2018Sparse}.

However, existing studies are typically limited to investigating single network architectures, lacking systematic comparison between different mainstream deep learning frameworks \cite{zhang2022towards, hu2024fast,tang2024super,Li2024Wind}. This absence of comprehensive architectural evaluation represents a significant research gap, as different network structures may offer distinct advantages for capturing various aspects of flow physics and spatial relationships, necessitating the selection of appropriate architectures based on specific practical conditions and desired requirements. Furthermore, systematic evaluation of these methods using realistic experimental data remains limited, with most studies relying on simulation-based training that may not reflect real-world deployment conditions. \secondrevision{Moreover, most studies employ predefined sensor placement strategies (e.g., uniform grids) without data-driven optimization tailored to their specific datasets, whereas optimal placement can maximize information content while minimizing measurement cost \cite{manohar2018data, gao2023optimal}.}

\begin{figure*}[htbp]
    \centering
    \includegraphics[width=\textwidth]{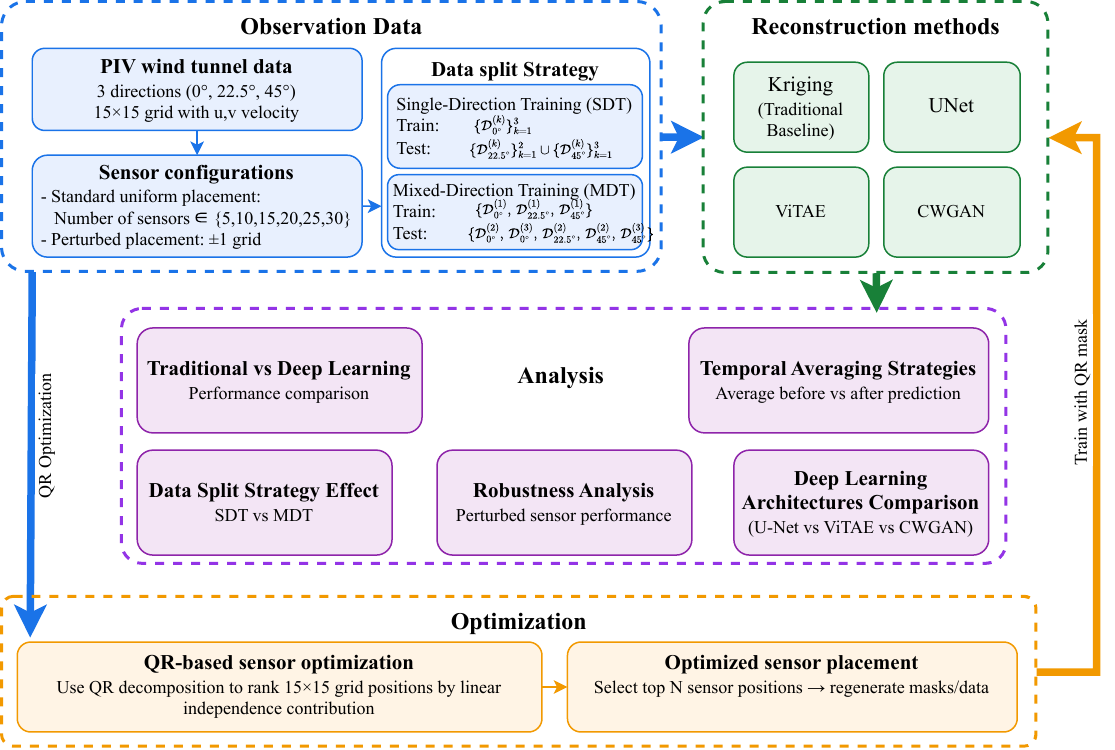}
    \caption{End-to-end workflow of real-time rooftop wind field reconstruction from PIV observations using sparse sensors, model comparison, and QR-based sensor optimization. For each wind direction $\theta \in \{0^\circ,\,22.5^\circ,\,45^\circ\}$, we collected $n_\theta$ realizations ($n_{0^\circ}=3$, $n_{22.5^\circ}=2$, $n_{45^\circ}=3$). Let $\mathcal{D}_{\theta}^{(k)}$ denote the $k$-th realization for direction $\theta$.}
    \label{fig:workflow}
\end{figure*}

This study addresses these limitations by establishing a learning-from-observation framework using high-resolution Particle Image Velocimetry (PIV) measurements of rooftop flow fields \secondrevision{and investigate the optimization of potential sensor placements.} By training reconstruction models directly on these experimental data, the framework enables a realistic evaluation of model performance under actual fluid conditions, accounting for noise, spatial resolution limits, and natural flow variability. The research objectives include: (1) quantitative comparison of deep learning methods (UNet, Vision Transformer Autoencoder, Conditional Wasserstein GAN) against traditional Kriging interpolation using experimental wind tunnel data; (2) evaluation of the robustness of methods under different sensor configurations and data splitting strategies; (3) investigation of multi-directional wind training strategies to enhance deep learning model generalization capabilities and their performance advantages relative to Kriging interpolation across different flow conditions; (4) development and validation of QR decomposition-based sensor position optimization methods; (5) analysis of temporal averaging strategies for data-limited scenarios. \fourthrevision{Through these investigations, the study provides methodological insight into model reliability under realistic experimental constraints, including limited training data, measurement noise, and turbulence variability, extending the contribution beyond case-specific benchmarking. In particular, the systematic comparison reveals the accuracy-complexity trade-offs of different architectures, which is essential for identifying appropriate models under practical constraints \cite{formont2025evaluation}. Specific findings address computational efficiency, robustness to sensor position uncertainties \cite{kaiser2024cluster}, and performance across varying sensor densities, collectively offering systematic guidance for method selection and sensor configuration in real-world applications.} Detailed workflow is shown in Fig. \ref{fig:workflow}.

The paper is organized as follows. Section 2 presents the methodology for four reconstruction approaches and their implementation details. Section 3 describes the experimental setup, including wind tunnel measurements, data splitting strategies, sensor configurations, and evaluation metrics. Section 4 presents results comparing deep learning methods with traditional approaches, analyzes performance differences among neural network architectures, evaluates training strategy effects, and assesses robustness to sensor perturbations. Section 5 introduces QR decomposition-based sensor optimization and validates its effectiveness. Section 6 concludes with key findings and future research directions.

\section{Methodology}

\fourthrevision{Four reconstruction methods are evaluated in this study: Kriging interpolation as a traditional baseline and three deep learning architectures, namely UNet, CWGAN, and ViTAE. These three architectures were selected because they represent distinct modeling philosophies: UNet addresses the reconstruction as a deterministic mapping using an encoder-decoder structure; CWGAN adopts a generative adversarial approach to capture nonlinear turbulent characteristics; and ViTAE combines Transformer attention mechanisms with convolutional processing for joint global and local feature extraction. This comparison reveals how different modeling approaches perform for rooftop wind field reconstruction under realistic experimental constraints.}


\subsection{Kriging Interpolation}

Kriging interpolation serves as the traditional baseline method for wind field reconstruction from sparse sensor measurements. As a geostatistical interpolation technique, Kriging provides optimal unbiased estimation through spatial correlation modeling, making it suitable for comparison with deep learning approaches in this study \cite{oliver1990kriging}.

The implementation employs ordinary Kriging with separate processing for the $u$ and $v$ velocity components. The method operates on the assumption of spatial stationarity, constructing optimal weights for sensor measurements based on their spatial relationships. Our implementation utilizes a Gaussian variogram model with correlation length optimized within the range of 0.5–10.0 grid cells (0.017H–0.35H), selected based on the 15$\times$15 rooftop domain characteristics. The nugget effect was set to zero while the sill parameter was normalized to 1.0 for consistent scaling. \fourthrevision{The nugget was set to zero to enforce exact interpolation, establishing Kriging as a strict deterministic baseline. This choice avoids introducing a uniform noise assumption across the heterogeneous rooftop flow field, where turbulence intensity varies considerably between separation bubbles, shear layers, and reattachment zones. Forcing Kriging to pass exactly through the noisy observations highlights the inherent limitations of spatial interpolation under realistic experimental conditions, which motivates the investigation of learning-based approaches.} The algorithm employs the PyKrige library and processing each sample independently.

While Kriging provides rigorous interpolation with uncertainty quantification, it exhibits limitations for complex flow field reconstruction. The spatial stationarity assumption becomes problematic in rooftop environments with flow separation and recirculation zones \cite{Risser2016Review:} . Additionally, the linear interpolation framework cannot capture nonlinear relationships characteristic of turbulent flow patterns, motivating the investigation of deep learning approaches \cite{Miao2024Interpolation}.

\subsection{UNet}

The UNet architecture is distinguished by its U-shaped structure, featuring a contracting path for context capture and an expansive path for precise localization. Originally developed for biomedical image segmentation, UNet demonstrates particular effectiveness in detailed reconstruction tasks using limited datasets \cite{ronneberger2015u}, making it potential for wind field reconstruction from sparse sensor measurements \cite{Nowak2024Optimisation}.

\begin{figure*}[htbp]
    \centering
    \includegraphics[width=\textwidth]{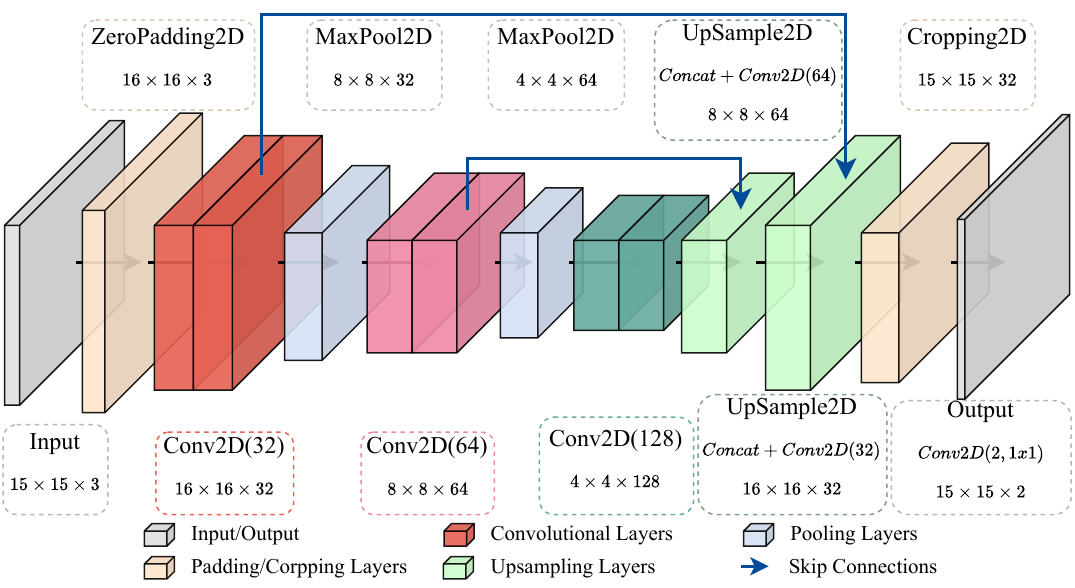}
    \caption{UNet architecture.}
    \label{fig:UNet}
\end{figure*}

Our UNet implementation is adapted to process wind field data with input dimensions of 15$\times$15$\times$3, accommodating the two velocity components and sensor mask information. The architecture employs zero-padding to transform the input from 15$\times$15 to 16$\times$16 for computational efficiency with standard convolution operations. The network progresses through a series of convolutional blocks where filter sizes gradually increase from 32 to 128 channels in the encoder pathway. \fourthrevision{The number of encoder blocks was determined by the spatial resolution of the input. With the 16$\times$16 input, three downsampling stages reduce the feature maps to 2$\times$2 at the bottleneck. A fourth stage would collapse the spatial dimension to 1$\times$1, destroying the spatial structure required for flow field reconstruction. The progressively widened filter configuration (32 to 128 channels) compensates for the limited depth, maintaining sufficient feature capacity while mitigating overfitting on the small dataset.} Each encoder block consists of two 3$\times$3 convolution layers with ReLU activation, followed by 2$\times$2 max-pooling for spatial downsampling. The decoder pathway employs upsampling operations to recover spatial resolution while progressively reducing feature depth (128→64→32 channels). Skip connections precisely merge feature maps from the contracting path with those in the expansive path, preserving vital spatial details throughout the network and ensuring both context and localization accuracy. The final 1$\times$1 convolution layer produces two-channel velocity field outputs, followed by cropping to restore the original 15$\times$15 spatial dimensions. This encoder-decoder architecture with skip connections enables UNet to effectively capture both global flow patterns and fine-scale spatial features essential for accurate wind field reconstruction. The detailed structure is visualized in Fig. \ref{fig:UNet}. The training strategy employs Adam optimization with adaptive learning rate reduction and early stopping mechanisms for optimal model generalization.

\subsection{Conditional Wasserstein GAN}

GAN represents an advanced deep learning approach \cite{goodfellow2014generativeadversarialnetworks} that addresses the limitations of traditional interpolation methods through adversarial training. The Conditional Wasserstein GAN framework extends the standard GAN architecture by incorporating conditional information from sparse sensor measurements and employing Wasserstein distance to enhance training stability \cite{arjovsky2017wasserstein, mirza2014conditionalgenerativeadversarialnets}.

\begin{figure*}[ht]
    \centering
    \includegraphics[width=\textwidth]{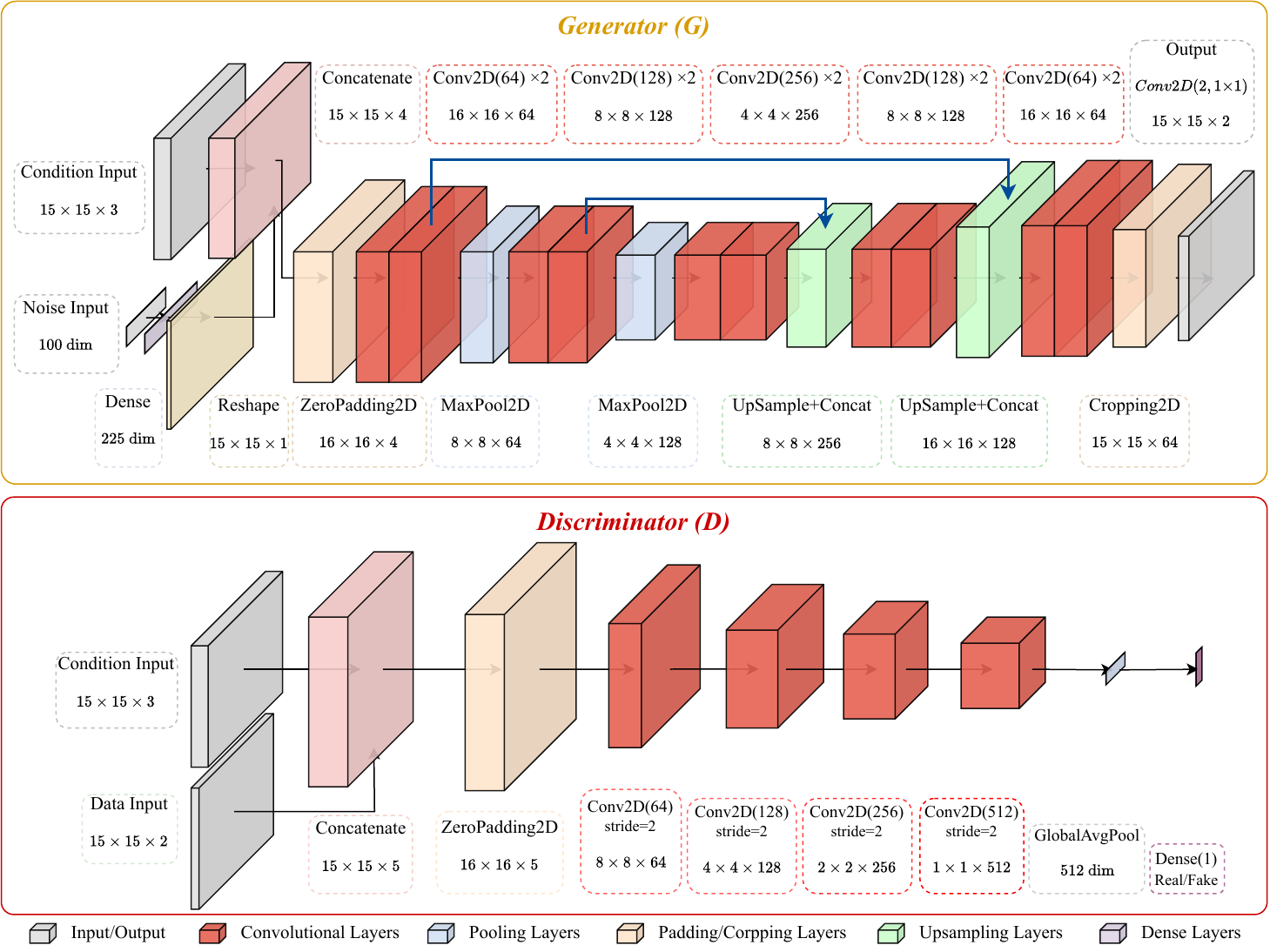}
    \caption{CWGAN architecture.}
    \label{fig:CWGAN}
\end{figure*}

The CWGAN architecture consists of two competing neural networks: a generator that reconstructs full wind fields from sparse sensor data, and a discriminator that distinguishes between reconstructed and ground truth flow fields. The generator employs a UNet-based architecture with encoder-decoder structure and skip connections, processing both sensor measurements and noise inputs. The network applies zero-padding to transform the 15$\times$15 input domain to 16$\times$16 for computational efficiency. The encoder pathway progressively reduces spatial resolution while increasing feature depth (64→128→256 channels), while the decoder pathway employs upsampling and skip connections to reconstruct spatial details. The discriminator employs strided convolutions with LeakyReLU activation and batch normalization, processing concatenated sensor conditions and flow field data through progressive feature extraction (64→128→256→512 channels). The network omits sigmoid activation to accommodate Wasserstein distance computation. The detailed structure is visualized in Fig. \ref{fig:CWGAN}. The training strategy employs Wasserstein distance for improved gradient stability. The generator loss combines adversarial objectives with L1 reconstruction loss (weighted 100:1) to balance realism and accuracy. Training uses five discriminator updates per generator update, Adam optimization (learning rate 0.0001), and early stopping for generalization. The conditional framework enables learning complex nonlinear mappings between sparse sensor measurements and full flow fields, particularly suitable for rooftop wind field reconstruction where traditional methods fail to capture turbulent characteristics.

\subsection{Vision Transformer Autoencoder}

Vision Transformer Autoencoder represents a hybrid approach that combines the global attention mechanisms of Transformer architectures with the spatial processing capabilities of convolutional neural networks \cite{dosovitskiy2020image,vaswani2023attentionneed}. Originally developed for computer vision tasks, ViTAE demonstrates effectiveness in capturing long-range dependencies and spatial relationships \cite{fan2025vitae}, making it particularly suitable for wind field reconstruction where flow patterns exhibit complex spatial correlations across the entire domain.

\begin{figure*}[htbp]
    \centering
    \includegraphics[width=\textwidth]{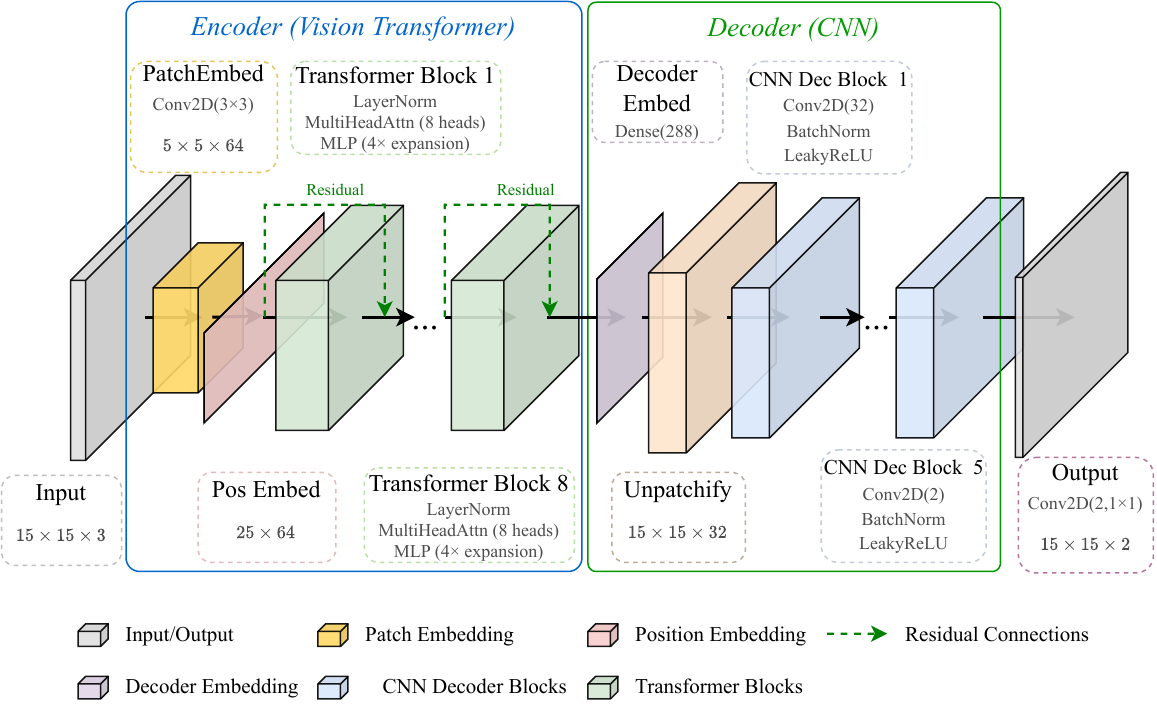}
    \caption{ViTAE architecture.}
    \label{fig:ViTAE}
\end{figure*}

Our ViTAE implementation employs a patch-based encoding strategy \cite{Xu2021ViTAE:} that divides the input wind field into non-overlapping patches for Transformer processing. The architecture processes input dimensions of 15$\times$15$\times$3 through patch embedding with 3$\times$3 patch size, creating a sequence of 25 patches that are linearly projected to 64-dimensional feature vectors. Learnable 2D sinusoidal position encodings are added to preserve spatial relationships between patches, enabling the model to understand the geometric structure of the wind field data. The encoder consists of eight Transformer blocks, each containing multi-head self-attention mechanisms with eight attention heads and Multi-Layer Perceptron (MLP) layers with expansion ratio of 4.0. The self-attention mechanism enables the model to capture global dependencies between different spatial regions, while layer normalization and residual connections ensure stable training dynamics. Following the Transformer encoder, a CNN decoder reconstructs the spatial wind field through a series of convolutional blocks that progressively refine the feature representations to produce two-channel velocity field outputs. The detailed structure is visualized in Fig. \ref{fig:ViTAE}. The training configuration follows similar optimization and regularization strategies as the previous methods. This hybrid architecture enables ViTAE to learn both global flow patterns through attention mechanisms and fine-scale spatial features through convolutional processing, effectively bridging the gap between global context understanding and local feature extraction for accurate wind field reconstruction.


\section{Experimental Setup}

\subsection{Wind Tunnel Experiment and Dataset}
\begin{figure*}[ht]
    \centering
    \includegraphics[width=0.7\textwidth]{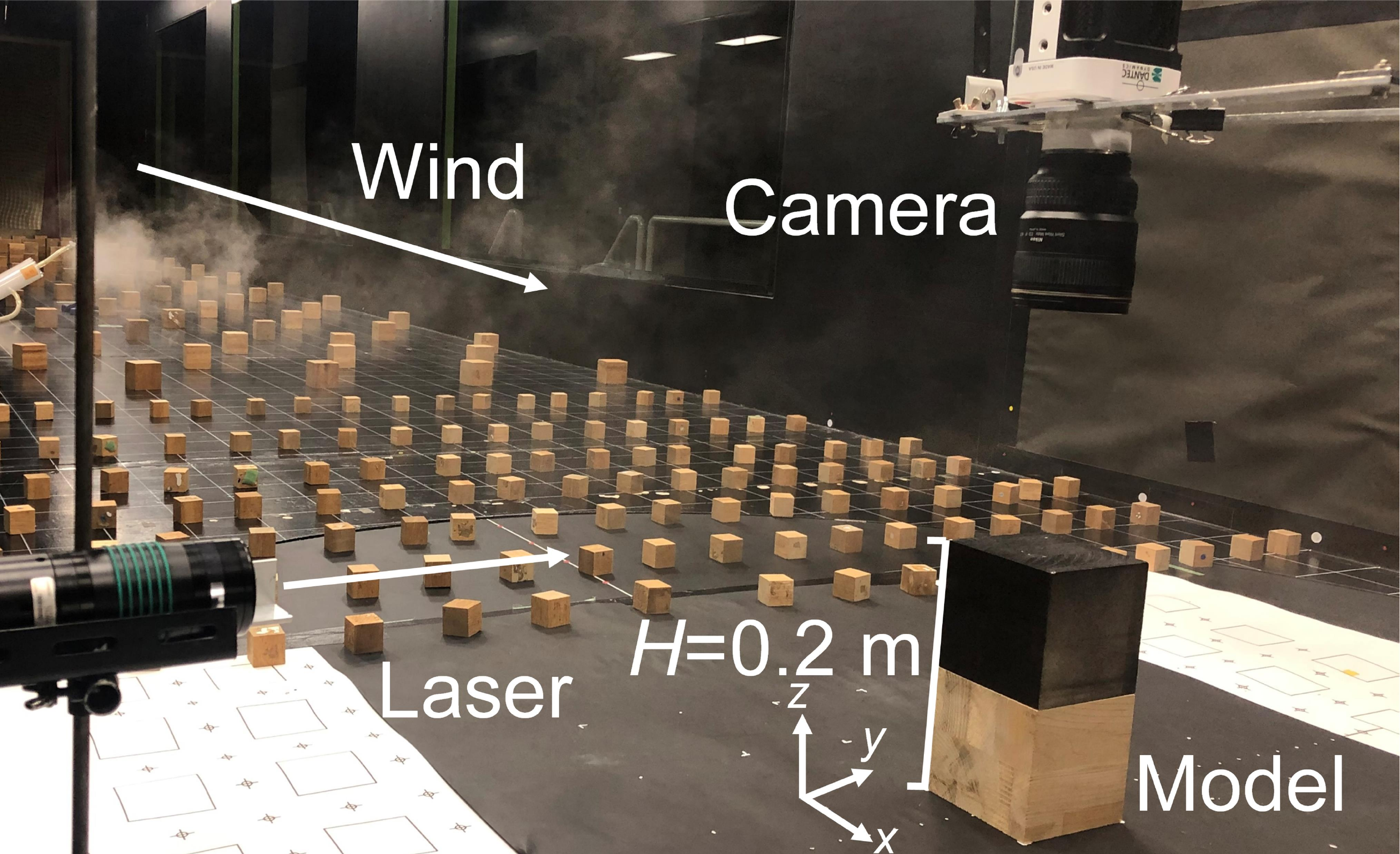}
    \caption{Wind tunnel experiment layout.}
    \label{fig:WTE_setting}
\end{figure*}

Experiments were conducted in the boundary layer wind tunnel at the Institute of Industrial Science, University of Tokyo \cite{lin2025wind}. The experimental layout is illustrated in Figure~\ref{fig:WTE_setting}. The model was a single rectangular block with a height-to-width-to-length ratio of 1:1:2. It was built at a geometric scale of 1:200, corresponding to a \thirdrevision{model} height of $H = 0.2$~m. The model was then placed in a simulated turbulent boundary layer. Tests were performed for three approach wind directions: $0^\circ$, $22.5^\circ$, and $45^\circ$. The $x$, $y$, and $z$ axes represent the streamwise, spanwise, and vertical coordinates, respectively.

The incoming flow profile followed the power law with an exponent of 0.22, and the reference velocity at the model height, $U_H$, was maintained at 0.70~m/s. Instantaneous horizontal velocity fields at $z/H = 1.05$ were obtained using PIV, enabling spatially continuous measurements with high temporal and spatial resolution. The temporal resolution was 0.001~s, and the spatial resolution was $0.035\,H$ in both the $x$ and $y$ directions. Each run yielded 8~s of data. For analysis, three datasets were collected for the $0^\circ$ and $45^\circ$ cases, and two for the $22.5^\circ$ case. The instantaneous velocity components in the $x$ and $y$ directions are denoted by $u$ and $v$, respectively. All velocities were normalized by $U_H$, and the discussion is based on these non-dimensional values. The normalized horizontal wind speed is defined as $w_s = \sqrt{u^2 + v^2} / U_H$, with its time-averaged value and standard deviation denoted by $WS$ and $ws_{SD}$.

\begin{figure*}[htbp]
    \centering
    \includegraphics[width=0.9\textwidth]{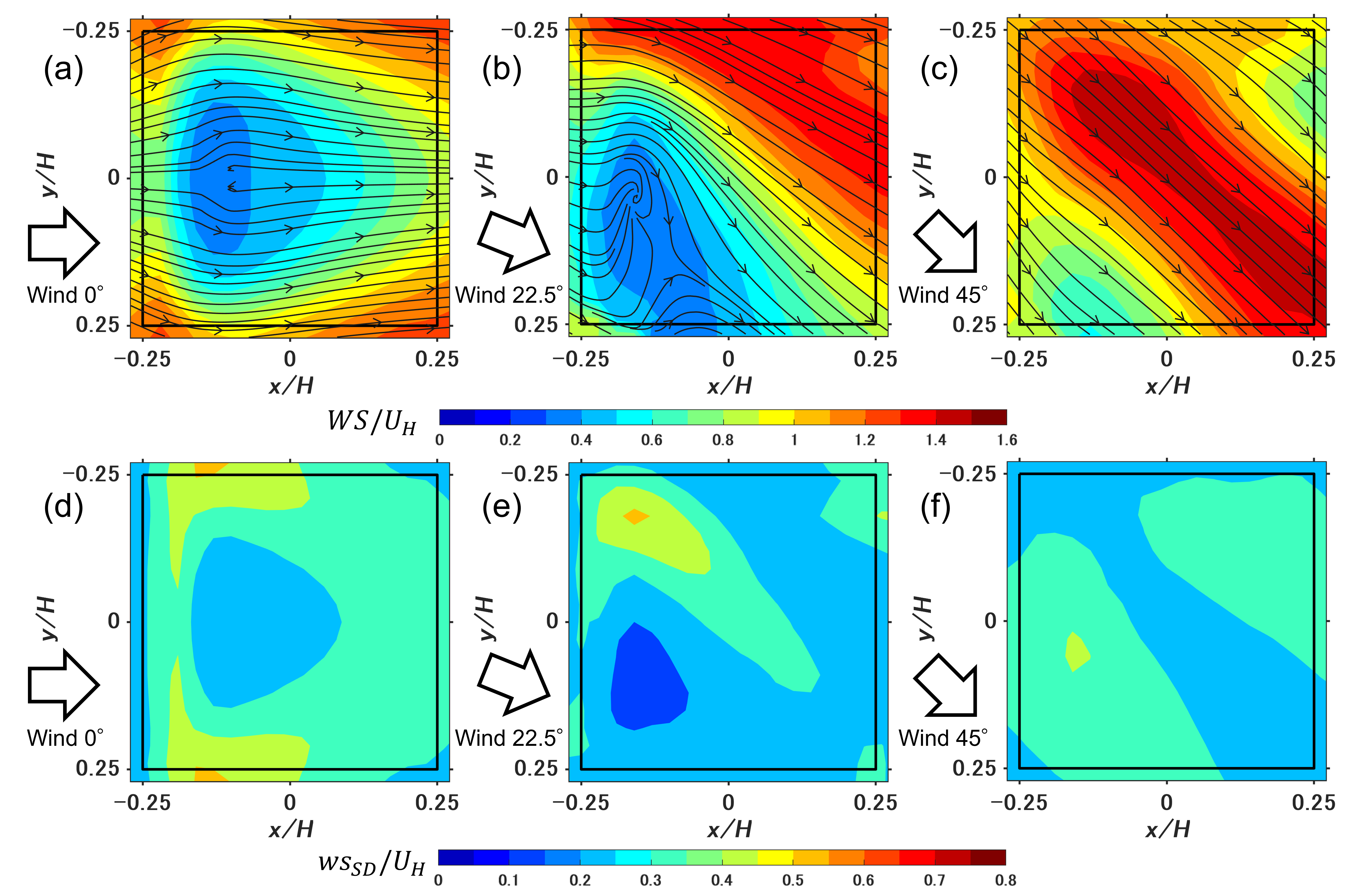}
    \caption{Distributions of time-averaged value and standard deviation of horizontal wind velocity in $z$/$H$ = 1.05 under different wind directions.}
    \label{fig:WTE_Results}
\end{figure*}

Figure~\ref{fig:WTE_Results} shows Distributions of time-averaged value and standard deviation of horizontal wind velocity under different wind directions. For mean horizontal wind speed, under the 0-degree wind direction (Figure~\ref{fig:WTE_Results} (a)), the rooftop flow field is characterized by the flow separation bubble at the building frontal edge and the reattachment at the middle of the roof with a low $WS$ region at the upwind half of the roof. The 22.5-degree wind direction induces asymmetric conical vortices, resulting in an uneven spatial distribution of high- and low-speed zone of $WS$ across the rooftop (Figure~\ref{fig:WTE_Results} (b)). Under the 45-degree wind direction (Figure~\ref{fig:WTE_Results} (c)), the flow field is characterized by two counter-rotating conical vortices that form along the roof’s diagonal axis. In detail, high-speed zone of $WS$ is found along the roof’s diagonal axis because of the channelling effect while two low $WS$ regions appear near the building corners due to corner vortices.

For standard deviation of horizontal wind velocity, under the 0-degree wind direction (Figure~\ref{fig:WTE_Results} (d)), high $ws_{SD}$ is found on the rooftop compared to surrounding locations because of the flow separation, and low $ws_{SD}$ is found in the flow separation bubble at the roof center. Under the 22.5-degree wind direction (Figure~\ref{fig:WTE_Results} (e)), high $ws_{SD}$ is found near the building frontal corner of the building and low $ws_{SD}$ is found in the recirculation region. Under the 45-degree wind direction (Figure~\ref{fig:WTE_Results} (f)), high $ws_{SD}$ is found in two stagnation regions and low $ws_{SD}$ is found along the roof’s diagonal axis.

For each wind direction $\theta \in \{0^\circ,\,22.5^\circ,\,45^\circ\}$, we collected $n_\theta$ realizations ($n_{0^\circ}=3$, $n_{22.5^\circ}=2$, $n_{45^\circ}=3$). Let $\mathcal{D}_{\theta}^{(k)}$ denote the $k$-th realization for direction $\theta$. \fourthrevision{Each realization comprises $7.999$~s of time-resolved PIV data sampled at $\Delta t=0.001$~s, yielding $7{,}999$ temporal snapshots. This study focuses on evaluating reconstruction approaches under site-specific experimental constraints, reflecting the practical scenario in which rooftop wind models are typically trained and deployed for a target building. In practical applications, flow characteristics are strongly site-dependent, and the current dataset represents a realistic deployment scenario. Given the high sampling rate, these snapshots provide statistically independent flow realizations due to turbulence variability. The results demonstrate that even with data from a single measurement plane, the evaluated methods can effectively reconstruct the velocity field. This is relevant to experimental fluid mechanics, where instantaneous PIV datasets are inherently limited in availability.}

\subsection{Data Splitting Strategies}

As the rooftop wind-speed distribution strongly depends on the approaching wind direction, this raises the question of whether a model trained on a single wind direction can accurately reconstruct flow fields under different wind directions. Therefore, two distinct data splitting strategies as shown in the workflow Fig. \ref{fig:workflow} are employed to evaluate model performance under different training scenarios and assess generalization capabilities across multiple wind directions. These strategies examine the effectiveness of single-direction versus multi-directional training approaches for wind field reconstruction tasks.

Under the SDT split, models are trained exclusively on the 0° realizations $\{\mathcal{D}_{0^\circ}^{(k)}\}_{k=1}^{3}$ and evaluated on realizations from other wind directions $\{\mathcal{D}_{22.5^\circ}^{(k)}\}_{k=1}^{2} \cup \{\mathcal{D}_{45^\circ}^{(k)}\}_{k=1}^{3}$. This setting assesses cross-direction generalization from a single wind direction to unseen flow patterns. Under the MDT split, the training set contains one realization from each direction $\{\mathcal{D}_{0^\circ}^{(1)},\,\mathcal{D}_{22.5^\circ}^{(1)},\,\mathcal{D}_{45^\circ}^{(1)}\}$, and testing uses the remaining \thirdrevision{independent} realizations across all directions $\{\mathcal{D}_{0^\circ}^{(2)},\,\mathcal{D}_{0^\circ}^{(3)},\,\mathcal{D}_{22.5^\circ}^{(2)},\,\mathcal{D}_{45^\circ}^{(2)},\,\mathcal{D}_{45^\circ}^{(3)}\}$. \thirdrevision{Crucially, the data splitting is performed by independent experimental realizations rather than random snapshot sampling. Since each realization represents a separate experimental session, there is no temporal continuity or overlap between the training and testing sets, thereby preventing temporal leakage. Besides, robustness analyses presented in \ref{appendix:mdt_robustness} verify that realizations within the same wind direction exhibit high structural similarity (average SSIM > 0.95) while maintaining sufficient independence for model evaluation, and that model performance is insensitive to the particular choice of training run (variations mostly within $\pm$2\% for SSIM and FAC2).}

\thirdrevision{For both strategies, the designated training realizations are further partitioned into training and validation subsets using an 80-20 split with random state 42 for reproducibility. All deep learning models employ validation-based early stopping with a patience of 20 epochs, monitoring validation loss to prevent overfitting. The combination of validation set partitioning and early stopping ensures that models generalize well to unseen data rather than memorizing training samples.}



\subsection{Sensor Configurations}

Six sensor quantity configurations (5, 10, 15, 20, 25, 30) are evaluated to assess reconstruction performance under varying spatial coverage densities. For each sensor quantity, the baseline sensor positions are determined using a grid-based uniform distribution strategy. The spatial domain is systematically divided into regions based on Voronoi tessellation \cite{wang2024dynamical, cheng2024efficient, fukami2021global}, with sensors placed at the geometric center of each region to ensure uniform spatial coverage across the 15$\times$15 grid domain.

\begin{figure*}[htbp]
    \centering
    \includegraphics[width=\textwidth]{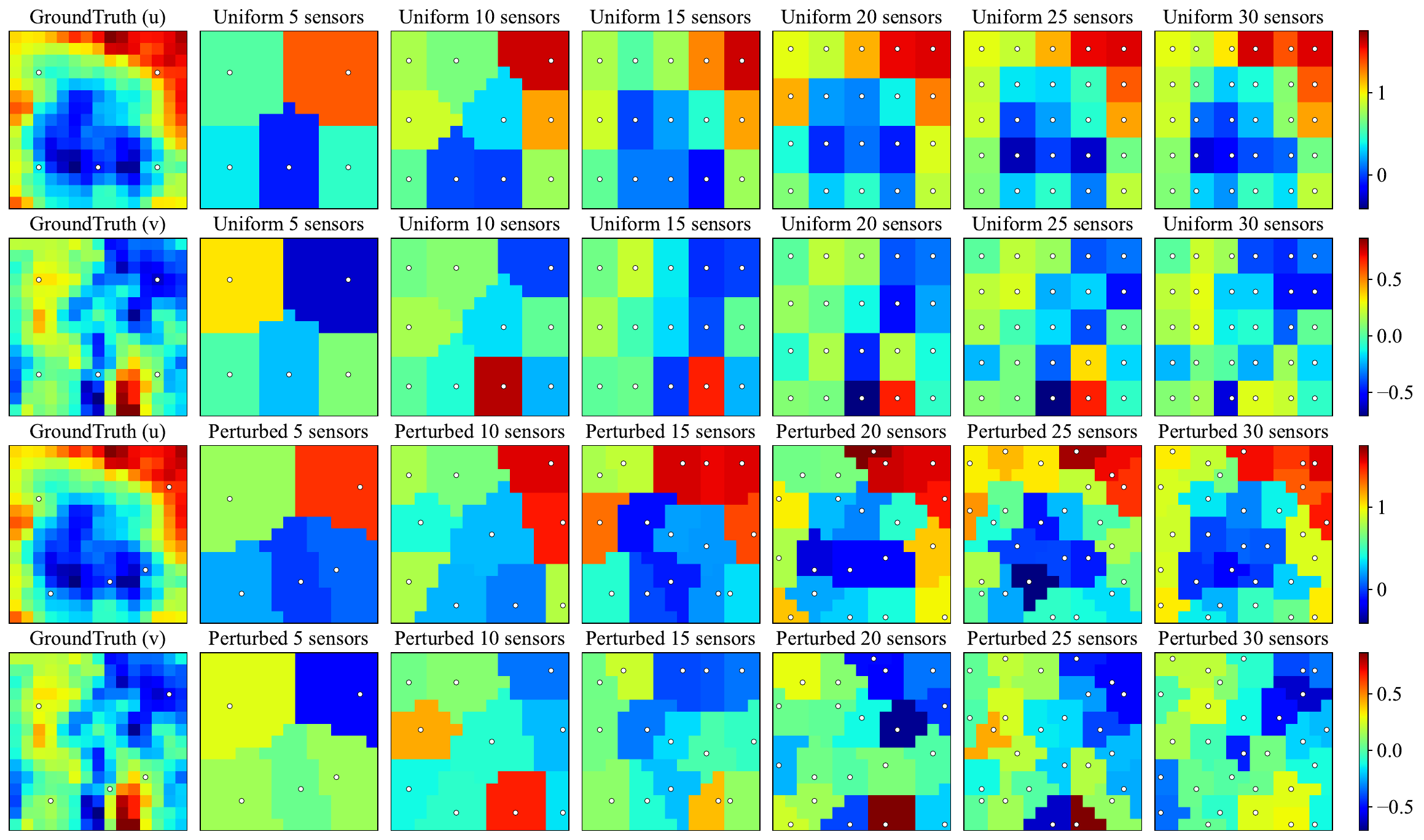}
    \caption{Ground truth, uniform sensor placement and perturbed sensor placement examples for different sensor quantities, white dots represent the sensor positions. \secondrevision{Polygonal boundaries show Voronoi cells, each cell contains all points closest to that cell's sensor (white dot), ensuring uniform spatial coverage.}}
    \label{fig:dataset}
\end{figure*}

To evaluate model robustness under realistic deployment conditions, \firstrevision{such as sensor position offsets or sensor failures caused by extreme weather conditions (eg. snow, rain, strong winds) \cite{gao2024uncertainties}}, perturbed sensor position datasets are generated for each sensor quantity configuration. Each sensor position experiences random perturbations within a range of ±1 grid cell in both horizontal and vertical directions, with the perturbed positions constrained to remain within the domain boundaries. The perturbation magnitude of ±1 grid cell is selected based on the 15$\times$15 grid resolution: larger perturbations (e.g., ±2 grid cells) would introduce excessive displacement relative to the domain size, while sub-grid perturbations are not applicable in the discrete grid framework. This perturbation approach simulates practical deployment scenarios where sensor positions may deviate from their intended placements. \secondrevision{Sensor positions differ across configurations, but within each configuration, positions remain identical between training and testing sets.} Figure~\ref{fig:dataset} shows the ground truth and sensor position examples for different sensor quantities.

To evaluate the effectiveness of the QR optimization, QR-optimized sensor positions are also implemented for each sensor quantity to compare with the uniform distribution approach. These optimal placements are determined through QR decomposition analysis of Proper Orthogonal Decomposition basis functions extracted from the training data, identifying sensor positions that maximize the linear independence of measurements for enhanced reconstruction performance. The detailed methodology and theoretical foundation for QR-based sensor optimization are presented in Section 5.

\subsection{Evaluation Metrics}

Four complementary evaluation metrics are employed to comprehensively assess wind field reconstruction performance across different aspects of prediction accuracy and reliability. \thirdrevision{All metrics (MG, NMSE, FAC2, and SSIM) are calculated separately for the streamwise velocity component ($u$) and spanwise velocity component ($v$), and then averaged. This component-wise evaluation approach allows independent assessment of prediction accuracy in both flow directions while providing a comprehensive overall performance measure.}

The Structural Similarity Index (SSIM) evaluates the perceptual quality of reconstructed velocity fields by comparing structural information between predicted and ground truth data. SSIM values range from 0 to 1, with higher values indicating better structural preservation.

The Geometric Mean Bias (MG) quantifies systematic bias in predictions by computing the exponential of the mean logarithmic \thirdrevision{ratio} between experimental and predicted values:
\thirdrevision{
    \begin{equation}
        MG = \exp\left(
        \frac{1}{N} \sum_{i=1}^{N}
        \ln\left(\frac{O_i}{P_i}\right)
        \right)
    \end{equation}}
\thirdrevision{where $O_i$ and $P_i$ denote the observed and predicted velocity component values (u or v) at location $i$, respectively. An ideal value of 1.0 indicates unbiased predictions, while values greater than 1.0 suggest overprediction and values less than 1.0 indicate underprediction. Since MG evaluates bias through logarithmic ratios, only data points where $O_i$ and $P_i$ have the same sign (i.e., $O_i \cdot P_i > 0$) are included in the calculation. This ensures that MG evaluates bias only when flow direction is correctly predicted, as assessing bias is physically meaningless when the predicted flow direction is opposite to the observed direction. This filtering criterion is applied consistently across all models and evaluation scenarios.}

The Normalized Mean Square Error (NMSE) provides a dimensionless measure of overall prediction accuracy by normalizing the mean square error with the product of experimental and predicted means:
\begin{equation}
    \text{NMSE} =
    \frac{
        \frac{1}{N}\sum_{i=1}^{N} (O_i - P_i)^2
    }{
        \left(\frac{1}{N}\sum_{i=1}^{N} O_i\right)
        \left(\frac{1}{N}\sum_{i=1}^{N} P_i\right)
    }
\end{equation}
\thirdrevision{where $O_i$ and $P_i$ denote the observed and predicted velocity component values (u or v) at location $i$, respectively.}
The numerator measures the mean squared error between predictions and observations, while
the denominator normalizes this error by the product of their mean magnitudes. Lower NMSE values indicate better reconstruction accuracy, with 0 representing perfect prediction. This normalization allows for comparison across different flow conditions and velocity magnitudes.

The Factor of 2 (FAC2) metric evaluates prediction reliability by calculating the fraction of data points that satisfy specific accuracy criteria. FAC2 is computed as:

\begin{equation}
    FAC2 = \frac{1}{N} \sum_{i=1}^{N} n_i
\end{equation}

where $n_i$ is defined as:

\begin{equation}
    n_i = \begin{cases}
        1, & \text{if } 0.5 \leq \frac{P_i}{O_i} \leq 2        \\
        1, & \text{if } |O_i| \leq W \text{ and } |P_i| \leq W \\
        0, & \text{otherwise}
    \end{cases}
\end{equation}

Here, \thirdrevision{ $P_i$ and $O_i$ represent the predicted and observed velocity component values (u or v) at location $i$, respectively, and $W$ is the error tolerance threshold (typically 0.005).} FAC2 values range from 0 to 1, with higher values indicating more reliable predictions. The dual-criteria design allows the metric to accommodate both relative accuracy assessment (through the ratio criterion) and absolute accuracy for small values (through the threshold criterion).

\section{Results and Discussion}

\subsection{Deep Learning vs Traditional Methods}

\begin{figure*}[ht]
    \centering
    \begin{subfigure}{\textwidth}
        \centering
        \includegraphics[width=\textwidth]{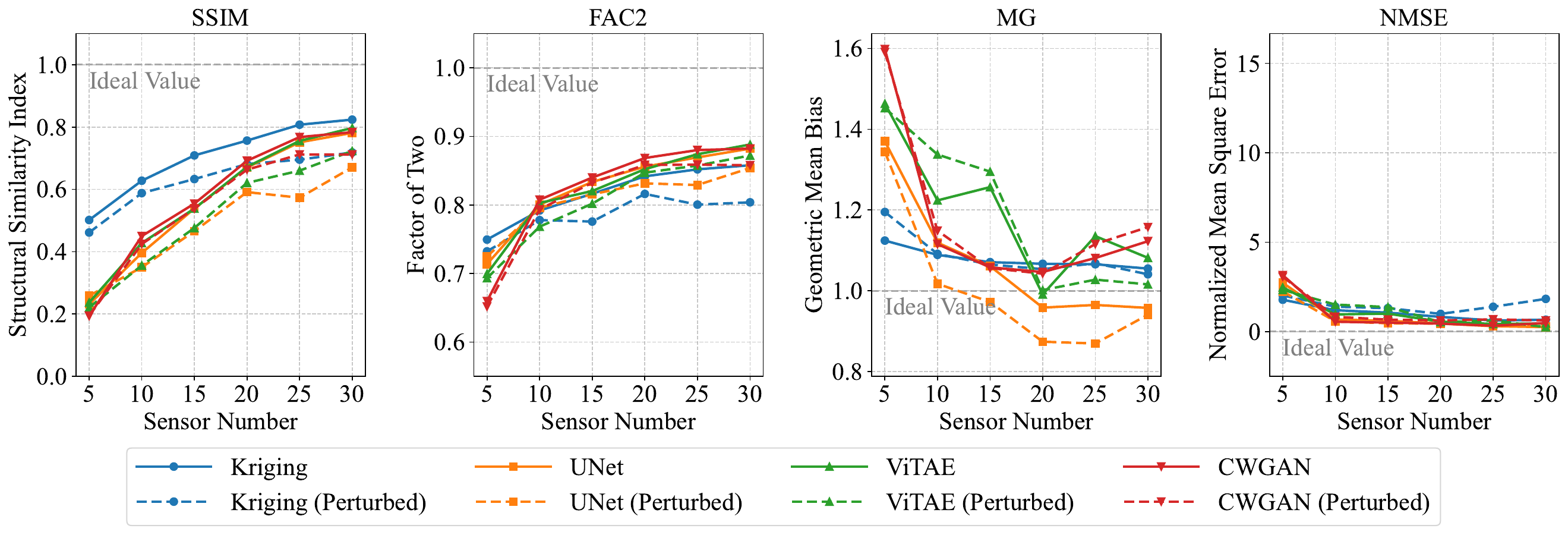}
        \caption{Comparison of metrics across models with standard sensor placement under SDT. Kriging outperforms deep learning at low sensor counts; with 5 sensors, Kriging achieves SSIM = 0.502 versus 0.194–0.237 for deep learning. As the number of sensors increases, deep learning gradually closes the gap in SSIM and outperforms Kriging in NMSE and FAC2.}
        \label{fig:standard:a}
    \end{subfigure}

    \vspace{1em}

    \begin{subfigure}{\textwidth}
        \centering
        \includegraphics[width=\textwidth]{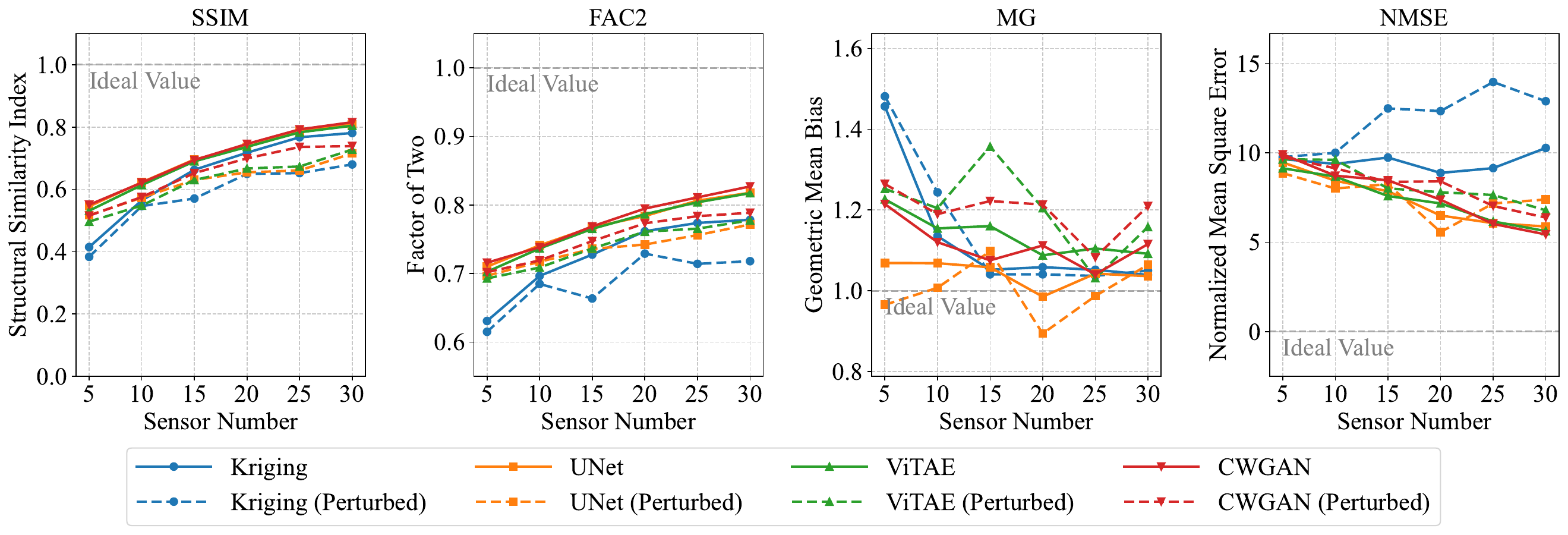}
        \caption{Comparison of metrics across models with standard sensor placement under MDT. Deep learning outperforms Kriging across most metrics and sensor counts. Comparing perturbed metrics further indicates higher robustness for deep learning relative to Kriging.}
        \label{fig:standard:b}
    \end{subfigure}

    \caption{Comparison of metrics across models with standard sensor placement under different training strategies.}
    \label{fig:standard}
\end{figure*}
\begin{figure*}[ht]
    \centering
    \includegraphics[width=\textwidth]{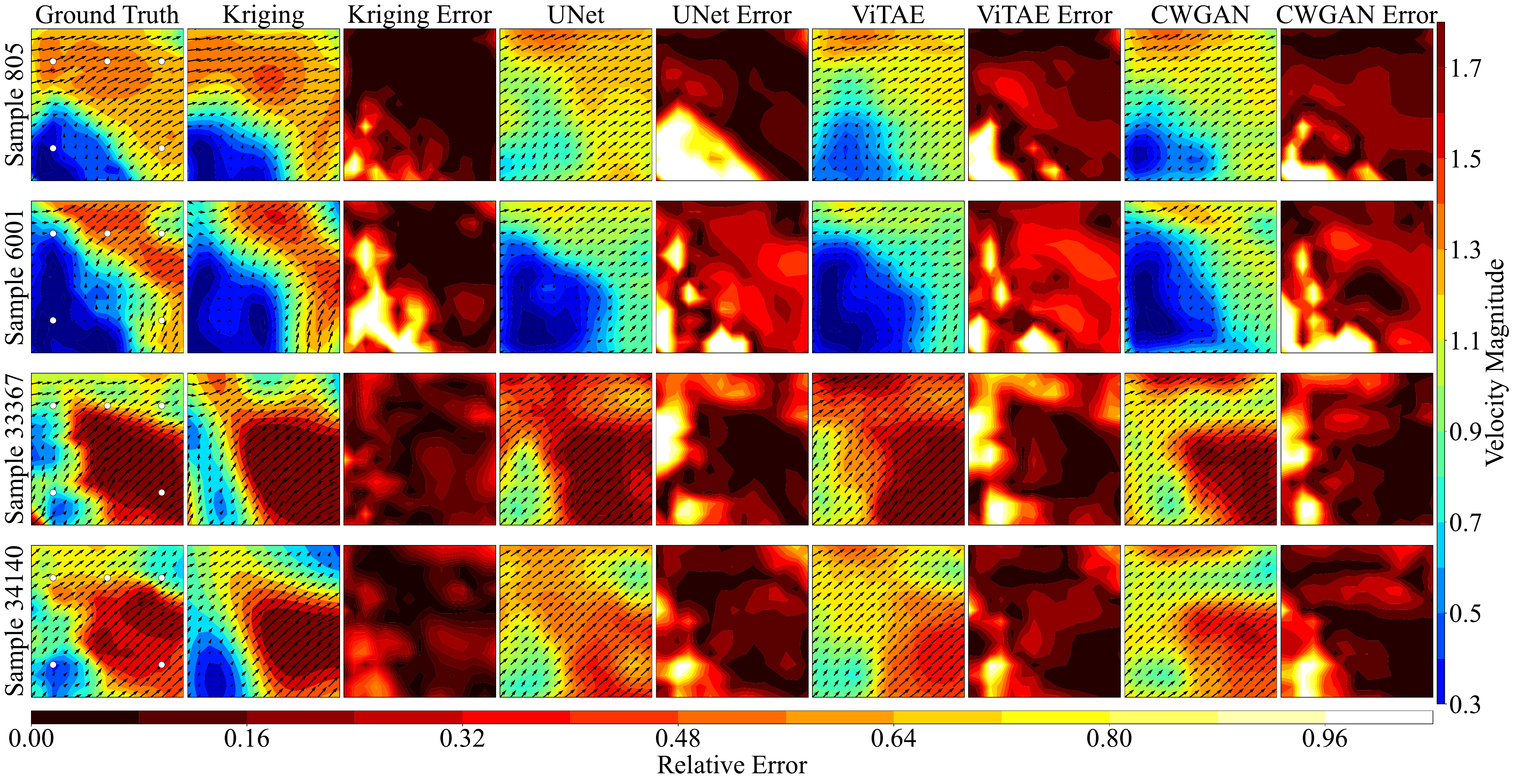}
    \caption{Visualization of predictions from 5 sensors under SDT (samples 805 and 6001 correspond to the 22.5° direction, whereas samples 33367 and 34140 correspond to 45° direction). Kriging outperforms deep learning methods at 5 sensors under SDT.}
    \label{fig:sensor_num_5_method_0}
\end{figure*}

The performance comparison between deep learning methods and traditional Kriging interpolation reveals complex, sensor-dependent patterns. Deep learning methods are significantly more efficient than Kriging, as shown in Table~\ref{tab:model_complexity}. The results, summarized in Fig.~\ref{fig:standard:a} and exemplified by Fig.~\ref{fig:sensor_num_5_method_0}, indicate that under SDT, Kriging demonstrated superiority in all metrics for sparse sensor configurations, achieving SSIM values of 0.502 and 0.628 for 5 and 10 sensors respectively, compared to deep learning methods which struggled at 0.194--0.237 and 0.396--0.450. This represents a 52.7--61.4\% performance gap for 5 sensors, suggesting that Kriging's spatial correlation assumptions are more effective when sensor coverage is limited and training set flow patterns are consistent.

A critical transition occurs at 20 sensors, where the performance dynamics shift. While Kriging maintains its SSIM advantage (0.756 vs. 0.670--0.692), deep learning methods achieve their most consistent NMSE superiority, with all three methods showing values of 0.450--0.552 compared to Kriging's 0.823. This 32.9--45.5\% improvement in NMSE indicates that deep learning methods increasingly leverage higher sensor density for overall error reduction.


The MG metric reveals unexpected behavior patterns. For sparse configurations (5 sensors), deep learning methods deviate severely from the ideal value of 1.0, reaching 1.370--1.598 compared to Kriging's 1.124 (22--42\% greater deviation). At 20 sensors, all deep learning methods achieve MG values closer to ideal than Kriging, with ViTAE showing the best performance (0.991, only 0.9\% deviation), followed by CWGAN (1.046, 4.6\% deviation) and UNet (0.958, 4.2\% deviation), compared to Kriging's 1.067 (6.7\% deviation). However, at higher sensor densities (25--30 sensors), only UNet maintains superior MG performance over Kriging, while ViTAE and CWGAN show increased bias deviation (8.2--13.5\% vs. Kriging's 5.5--6.6\%).

\begin{figure*}[ht]
    \centering
    \includegraphics[width=\textwidth]{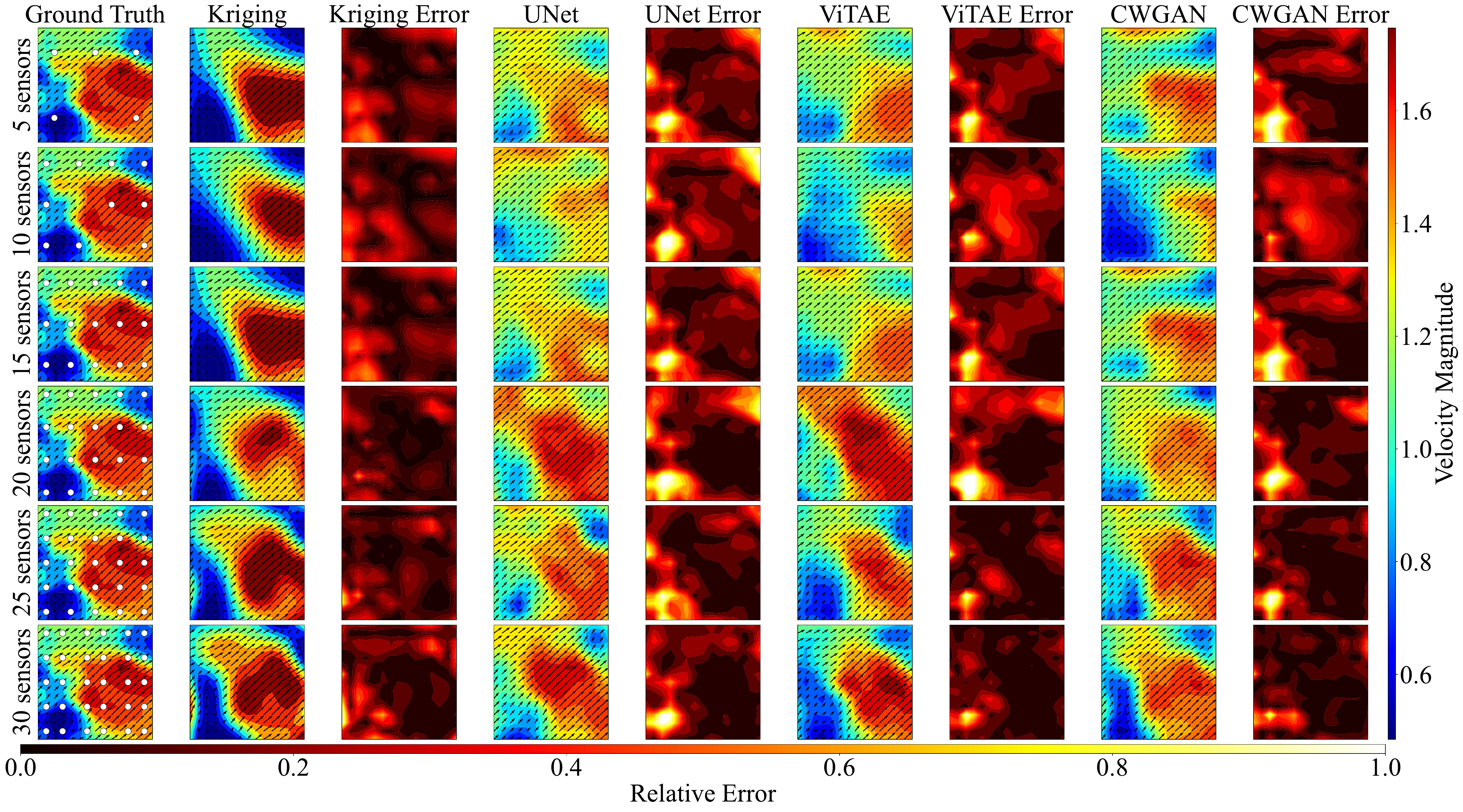}
    \caption{Prediction visualization across sensor counts under SDT (sample 34140 from 45° direction). As the sensor number increases, deep learning methods gradually close the gap in performance with Kriging.}
    \label{fig:sensor_num_comparison_method0}
\end{figure*}

The FAC2 metric provides complementary insights into prediction reliability. Under SDT, Kriging demonstrates superiority only at 5 sensors (0.749 vs. 0.660--0.714), but deep learning methods consistently outperform at higher sensor densities, achieving FAC2 values of 0.882--0.888 at 30 sensors compared to Kriging's 0.858. This pattern mirrors the NMSE transition at 20 sensors, reinforcing that deep learning methods become more reliable predictors as spatial information density increases, as evidenced by Fig. \ref{fig:sensor_num_comparison_method0}. The convergence of all deep learning methods toward FAC2 values above 0.85 at high sensor densities suggests that prediction reliability becomes less method-dependent when sufficient spatial coverage is available.

\begin{figure*}[ht]
    \centering
    \includegraphics[width=\textwidth]{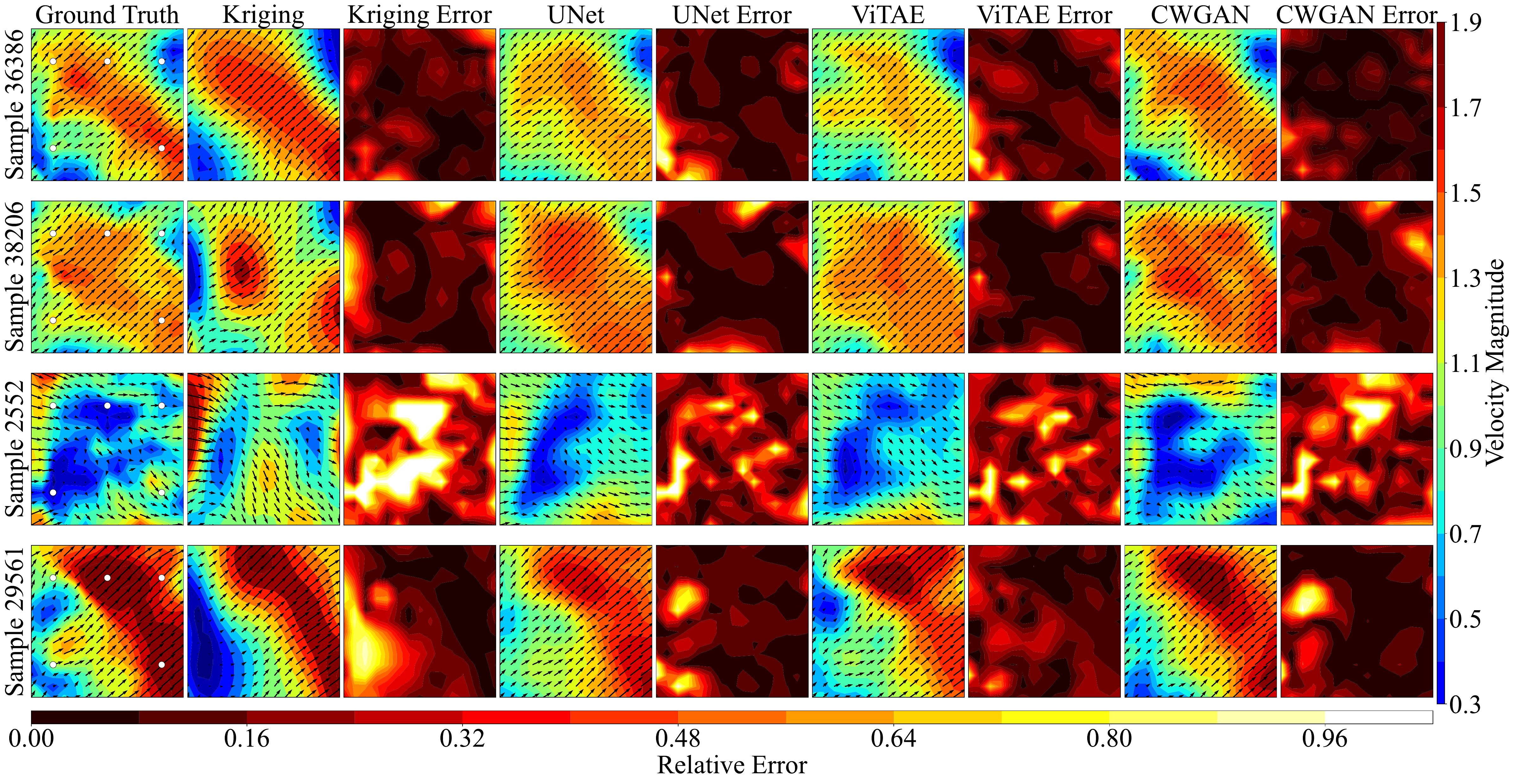}
    \caption{Sample 36386, 38206, and 29561 correspond to the 45° direction, whereas sample 2552 corresponds to the 0° direction. As described in Section 4.1, under MDT, deep learning methods generally outperform Kriging, particularly at 5 sensors. The figure shows predictions from 5 sensors under different wind directions training, demonstrating that deep learning methods provide more accurate predictions compared to Kriging. The predictions from Kriging deviate significantly from the ground truth, while the predictions from deep learning methods are closer to the ground truth.}
    \label{fig:sensor_num_5_method_1}
\end{figure*}

MDT fundamentally alters these relationships as shown in Fig.~\ref{fig:standard:b} and exemplified by Fig.~\ref{fig:sensor_num_5_method_1}. Deep learning methods consistently outperform Kriging in SSIM across all sensor configurations, with advantages ranging from 18.2--33.5\%. The most dramatic improvement occurs at 5 sensors (0.531--0.550 vs. 0.415), suggesting that exposure to diverse wind directions during training enables deep learning models to extract more robust features even from minimal sensor data. However, this comes at the cost of absolute performance: NMSE values are an order of magnitude higher (5.5--10.3) compared to SDT (0.25--3.1), indicating the inherent difficulty of reconstructing flows across multiple wind direction patterns. FAC2 analysis reinforces this advantage: all deep learning methods achieve values above 0.8 at 25 sensors (0.803--0.811), while Kriging remains consistently below this threshold throughout all configurations, reaching only 0.778 at 30 sensors. Regarding the MG metric, Kriging demonstrated superior performance with values closer to the ideal 1.0, while UNet achieved its optimal result of 0.958 at 20 sensors.


\begin{figure*}[ht]
    \centering
    \includegraphics[width=\textwidth]{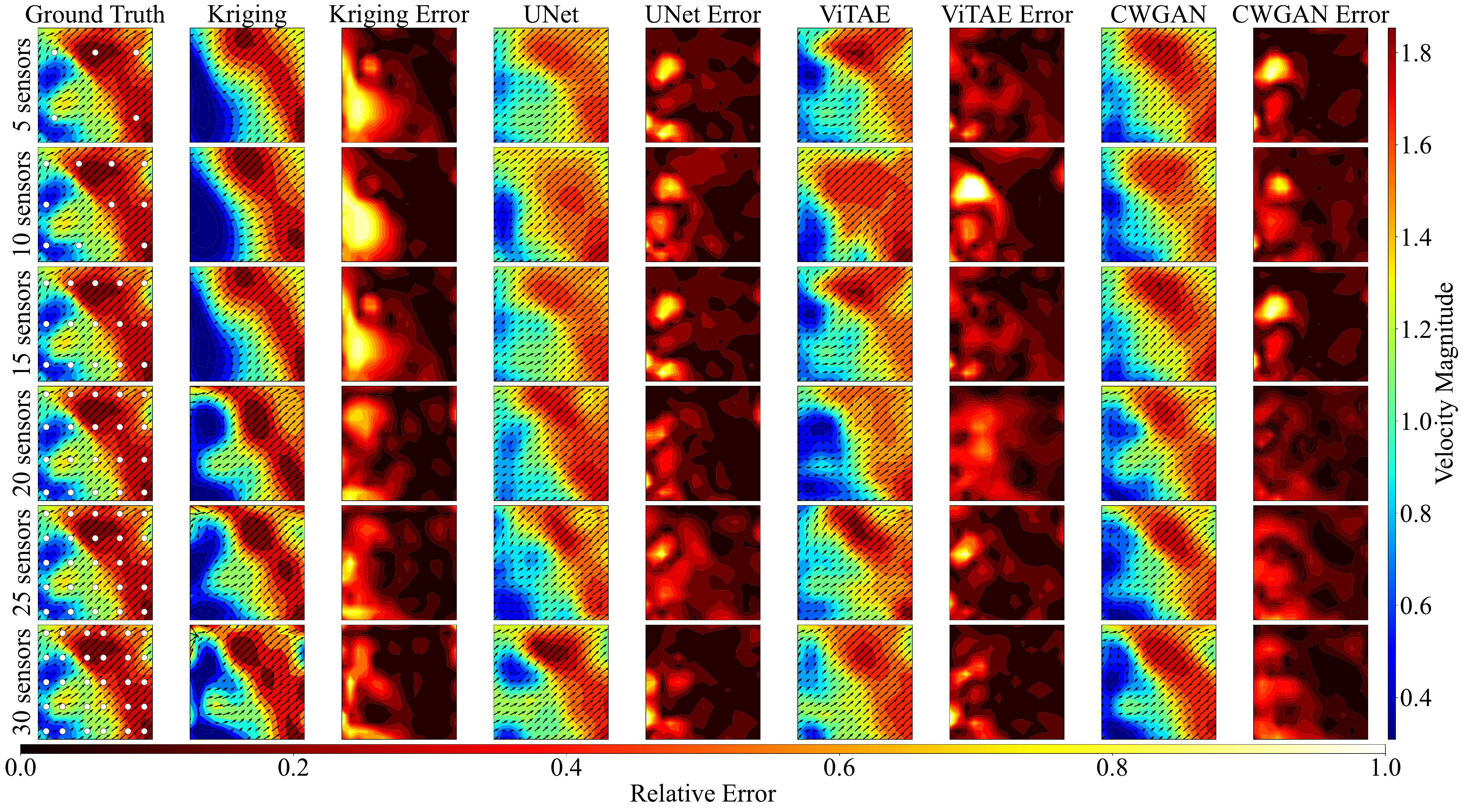}
    \caption{Prediction visualization across sensor counts under MDT (sample 29561 from 45° direction). For complex rooftop flow features (e.g., the top-left low-speed region), Kriging interpolation shows large deviations from the ground truth and fails to recover fine-scale structures, whereas deep learning methods better preserve local details. As the number of sensors increases, deep learning reconstructions converge toward the ground truth, consistent with the trends in Fig. \ref{fig:standard:b} (deep learning method performs better especially at sparse sensor configurations).}
    \label{fig:sensor_num_comparison_method1}
\end{figure*}

\subsection{Comparison among Deep Learning Methods}

The three deep learning architectures demonstrate distinct efficiency-performance profiles (Table~\ref{tab:model_complexity}). UNet and ViTAE have comparable parameter counts, yet ViTAE achieves 26.3\% lower computational cost (GFLOPs). CWGAN operates in a different complexity class with 18.7 times more parameters and 45.6 times more computation than the lightweight alternatives.

Architectural differences manifest in distinct performance patterns. UNet's encoder-decoder structure with skip connections demonstrates the highest geometric accuracy, maintaining MG values closest to ideal (0.986--1.072 under MDT) by preserving fine-scale spatial relationships. ViTAE's hybrid Transformer-CNN architecture achieves balanced but not exceptional performance, suggesting that its global attention mechanism may sacrifice some local geometric precision. CWGAN's adversarial training yields the highest SSIM (0.550--0.816 under MDT), but at the cost of geometric bias (MG: 1.018--1.218), indicating that the adversarial objective prioritizes perceptual quality over geometric accuracy.

Regarding training data diversity (detailed in Section~\ref{sec:data_splitting}), the three architectures reveal different learning mechanisms. UNet shows the most consistent improvement under MDT (24.4--131.5\% SSIM gain), suggesting that its hierarchical feature extraction is particularly effective at learning generalizable flow patterns. ViTAE shows more modest gains (20.8--124.4\%), indicating that its attention mechanism may require more data to fully exploit multi-directional diversity. CWGAN exhibits the most stable adaptation, with the smallest FAC2 degradation (5.7\% vs. 6.6\% for others), as its adversarial training provides a regularization effect that maintains reliability across heterogeneous training conditions.



For efficiency-performance trade-offs, ViTAE requires only 73.7\% of UNet's computation but achieves comparable SSIM under SDT (within 1.5\%), though UNet gains 0.5--2.8\% advantage under MDT. CWGAN's 4.7--9.4\% SSIM improvement comes at 45.6--62.0 times computational cost, yielding diminishing returns. However, CWGAN's better robustness to sensor perturbation (1.7--3.0\% FAC2 degradation vs. 3.6--4.4\% for others) suggests that the additional computational investment may be worthwhile for applications requiring high reliability under uncertain sensor conditions.

\begin{table}[ht]
    \centering
    \caption{\thirdrevision{Computational complexity and model characteristics of deep learning architectures. The inference time is normalized by the UNet baseline (approx. 0.109 ms per snapshot on an NVIDIA GeForce RTX 3080 Ti), and the deep learning methods show a significant advantage over Kriging.}}
    \label{tab:model_complexity}
    \begin{tabular}{lcccc}
        \toprule
        Model   & Parameters & GFLOPs & Model Size (MB) & Inference Time (per snapshot)                 \\
        \midrule
        UNet    & 471,586    & 0.0285 & 1.80            & \thirdrevision{1$\times$ ($\sim$0.109 ms)}    \\
        ViTAE   & 467,491    & 0.0210 & 1.78            & \thirdrevision{2.1$\times$ ($\sim$0.229 ms)}  \\
        CWGAN   & 8,770,000  & 1.301  & 33.46           & \thirdrevision{1.5$\times$ ($\sim$0.164 ms)}  \\
        Kriging & -          & -      & -               & \thirdrevision{13.7$\times$ ($\sim$1.493 ms)} \\
        \bottomrule
    \end{tabular}
\end{table}

\subsection{Effect of Data Splitting Strategies}
\label{sec:data_splitting}

Kriging and deep learning methods respond differently to training data diversity (solid lines in Fig.~\ref{fig:standard:a} and Fig.~\ref{fig:standard:b}). For Kriging, SDT consistently outperforms MDT, with SSIM advantages of 20.9--51.7\%, because its spatial stationarity assumption becomes invalid when mixing flow patterns from different wind directions. Deep learning methods show the opposite trend: MDT improves SSIM by 131.2--146.0\% at 5 sensors and maintains 4.0--7.4\% advantages at 30 sensors, with the benefit greatest for sparse configurations where models depend more on learned flow features. UNet shows the most consistent improvement (24.4--131.5\% in SSIM), while CWGAN achieves the highest absolute SSIM values under MDT (0.550--0.816).

The MG and FAC2 analyses reinforce these findings. UNet maintains near-ideal MG values (0.986--1.072) under MDT compared to significant deviations under SDT (0.958--1.370), while Kriging's MG degrades from near-ideal values under SDT (1.055--1.124) to substantial overestimation under MDT (1.039--1.457). For FAC2, Kriging suffers 11.0\% degradation from SDT to MDT, whereas deep learning methods show only 5.7--6.6\% degradation, with CWGAN being the most stable. This stability difference indicates that deep learning architectures are better for handling diverse training data.

The order-of-magnitude gap in NMSE between SDT (0.25--3.1) and MDT (5.5--10.3) is largely a normalization effect: mixing directions reduces per-sample mean magnitude, shrinking the denominator and inflating NMSE even when SSIM and FAC2 improve. Consequently, NMSE magnitudes should be compared within each split, while cross-split conclusions should rely on SSIM and FAC2.

\secondrevision{The significant performance drop of Kriging after switching from SDT to MDT is also worth analyzing. Spatial NMSE distribution by wind direction (Fig.~\ref{fig:spatial_comparison_SDT_5sensors} and Fig.~\ref{fig:spatial_comparison_MDT_5sensors}) reveals that 0° is particularly difficult to reconstruct due to its distinct flow pattern. Because the MDT test set contains 0° samples while SDT test set contains none, this discrepancy explains the observed degradation. Further analysis can be found in \ref{zero bad}.}

\thirdrevision{The superior performance of MDT raises the question of whether all wind directions are necessary in the training set. To address this, we conducted an additional experiment excluding 22.5° from training. The results (detailed in \ref{appendix:interpolation}) confirm that including all wind directions is essential for optimal performance.}


\subsection{Robustness to Sensor Perturbation}

Sensor position perturbation analysis reveals that robustness patterns are highly non-linear and method-dependent (dashed lines in Fig.~\ref{fig:standard:a} and Fig.~\ref{fig:standard:b}).

Under SDT, Kriging shows SSIM degradations of 5.4--13.9\%, peaking at 25 sensors. Deep learning methods exhibit more consistent responses: ViTAE shows 6.7--16.8\% degradation, while CWGAN achieves better robustness (3.3--8.2\% degradation), particularly at low sensor counts (0.4\% at 5 sensors) due to its adversarial training, though this advantage diminishes at higher densities. FAC2 degradations remain modest (1.7--4.1\%) for all methods under SDT, with CWGAN showing the best stability (1.7\%), indicating that basic prediction reliability within a factor of 2 remains intact despite SSIM degradation.

Under MDT, the pattern changes. FAC2 degradations increase to 3.0--5.5\%, with Kriging showing the highest sensitivity (5.5\%) due to more complex flow patterns amplifying spatial uncertainties. UNet shows notable NMSE stability (-4.8\% to 25.8\%), with instances where perturbed performance improves when slight position variations break symmetries in sensor configuration. ViTAE shows the highest sensitivity, with NMSE degradations reaching 81.5\% at 15 sensors, indicating that its transformer-based architecture depends more on precise spatial relationships.

Vulnerability peaks at intermediate densities (15--25 sensors) for most methods: sparse configurations have limited interpolation assumptions to violate, while dense configurations provide sufficient redundancy.

\subsection{Temporal Averaging Strategies for Data-Limited Scenarios}

\begin{figure*}[ht]
    \centering
    \begin{subfigure}{\textwidth}
        \centering
        \includegraphics[width=\textwidth]{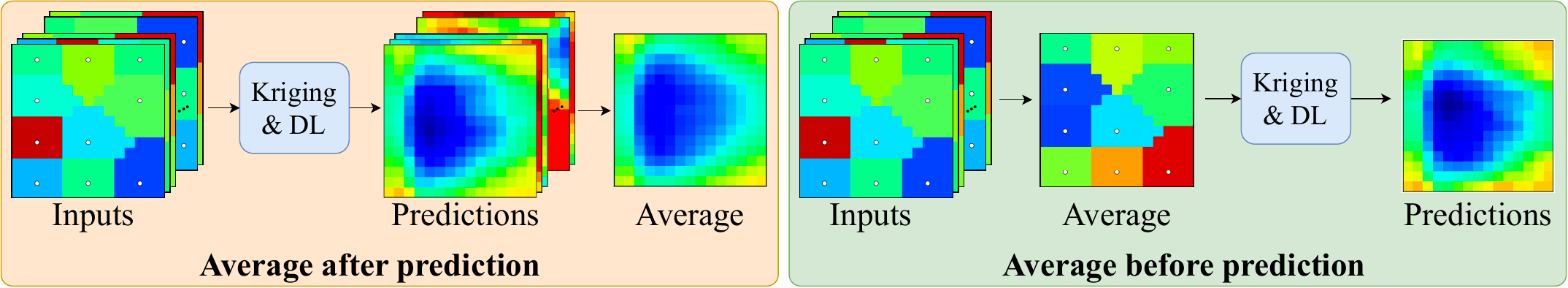}
        \caption{The two different pipelines of temporal averaging strategies. Average-after-prediction: individual wind fields are reconstructed first, then averaged to produce the final field. Average-before-prediction: multiple wind fields are averaged first, then the averaged field is reconstructed.}
        \label{fig:temporal_averaging:a}
    \end{subfigure}

    \vspace{1em}

    \begin{subfigure}{\textwidth}
        \centering
        \includegraphics[width=\textwidth]{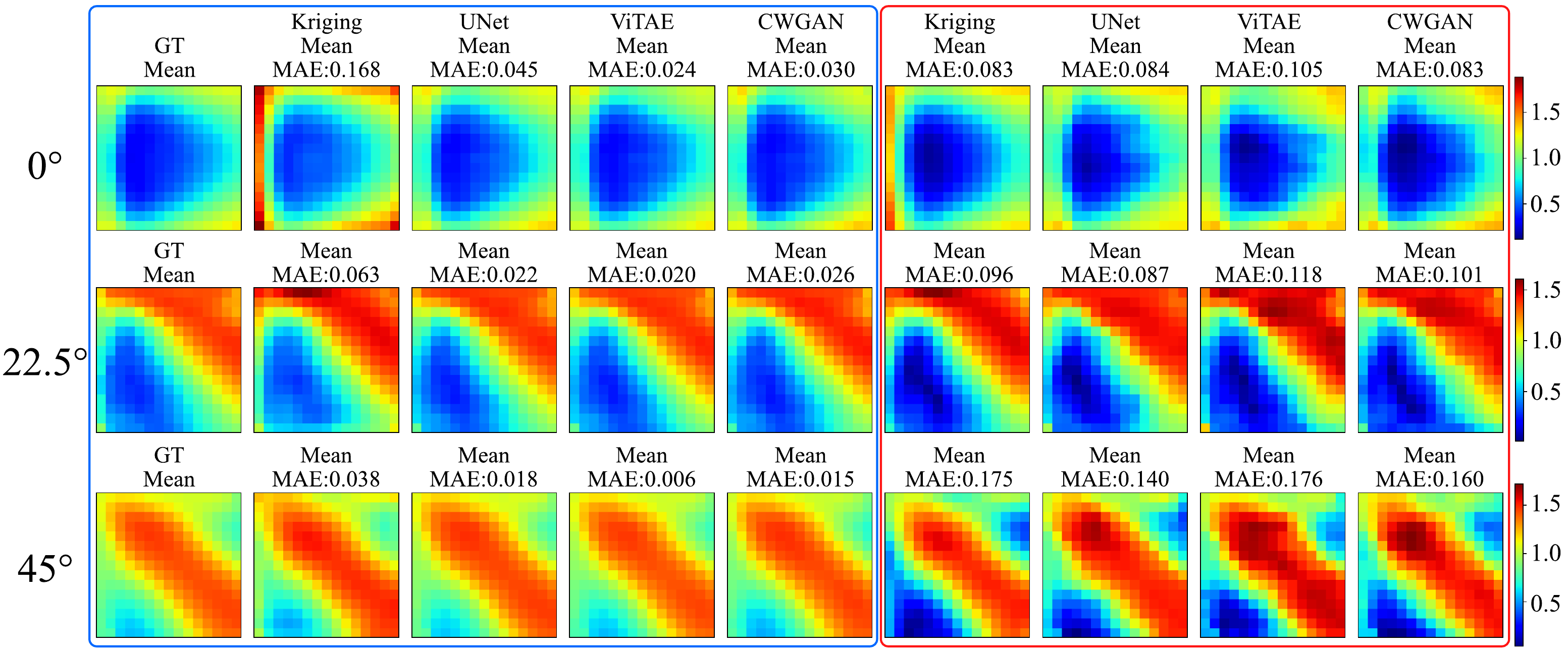}
        \caption{\firstrevision{Temporal averaging comparison between pre-averaging and post-averaging for MDT. Figures in the blue box represent the results of post-averaging, where mean and standard deviation are calculated temporally across multiple fields. Figures in the red box represent the results of pre-averaging, where errors are calculated spatially since only one averaged field is reconstructed. Overall, post-averaging performs better than pre-averaging, but exceptions exist. For example, pre-averaging Kriging outperforms post-averaging for the 0° wind direction. SDT has the same pattern as MDT.}}
        \label{fig:temporal_averaging:b}
    \end{subfigure}

    \caption{Temporal averaging strategies for data-limited scenarios.}
    \label{fig:temporal_averaging}
\end{figure*}

\firstrevision{In practical applications, the availability of temporal data may be limited due to measurement constraints or computational costs. To evaluate whether temporal averaging can be performed before reconstruction (pre-averaging) instead of after (post-averaging) as illustrated in Fig.~\ref{fig:temporal_averaging:a}, we compared the performance of both strategies across SDT, MDT, and sensor configurations.}

\firstrevision{Pre-averaging generally produces slightly lower performance compared to post-averaging for most methods and sensor densities (Fig.~\ref{fig:temporal_averaging:b}), though notable exceptions exist. Despite this, pre-averaging offers a practical advantage when temporal data are limited, as it requires only a single averaged field for reconstruction rather than multiple snapshots, reducing computational cost and data requirements. Given the relatively small performance difference, pre-averaging can serve as a viable alternative in data-constrained scenarios.}

\section{Sensor Position Optimization}

\subsection{QR-based Sensor Selection}

The QR decomposition method was employed to systematically identify optimal sensor positions based on Proper Orthogonal Decomposition (POD) analysis of the wind field dataset \cite{manohar2018data, gao2023optimal,xiao2019reduced}. This approach provides a mathematically rigorous framework for sensor placement optimization by leveraging the dominant flow structures captured in POD modes.

The POD basis construction begins with the wind field data matrix $\bY \in \mathbb{R}^{N \times 450}$, where $N$ represents the number of temporal snapshots and 450 corresponds to the spatial degrees of freedom (15$\times$15$\times$2 for $u$ and $v$ velocity components). The data matrix is centered by subtracting the temporal mean, and Singular Value Decomposition (SVD) is applied to extract the dominant flow modes:

\begin{equation}
    \bY_{centered} = \bU \bS \bV^T
    \label{eq:svd_pod}
\end{equation}

where $\bU$ contains the POD spatial modes, $\bS$ represents singular values, and $\bV^T$ contains temporal coefficients. The first \thirdrevision{$r=40$ POD modes are retained, accounting for approximately 84.6\% of the total flow energy}, forming the reduced basis matrix $\bPsi_r \in \mathbb{R}^{450 \times r}$.

The sensor selection algorithm employs QR decomposition with column pivoting on the transposed POD basis:

\begin{equation}
    \bPsi_r^T \bP = \bQ \bR
    \label{eq:qr_decomposition}
\end{equation}

where $\bP$ is the permutation matrix that reorders columns according to their importance, $\bQ$ is orthogonal, and $\bR$ is upper triangular. The permutation vector $\mathbf{p}$ obtained from $\bP$ directly provides the sensor importance ranking, with $p_1$ indicating the most informative sensor location.

This QR-based approach ensures that selected sensors maximize the linear independence of the measurement matrix $\bH \bPsi_r$, where $\bH$ represents the observation operator. The method systematically identifies sensor positions that provide the most linearly independent information about the dominant flow structures, thereby optimizing the reconstruction capability for a given number of sensors.

\begin{figure}[htbp]
    \centering
    \includegraphics[width=0.48\textwidth]{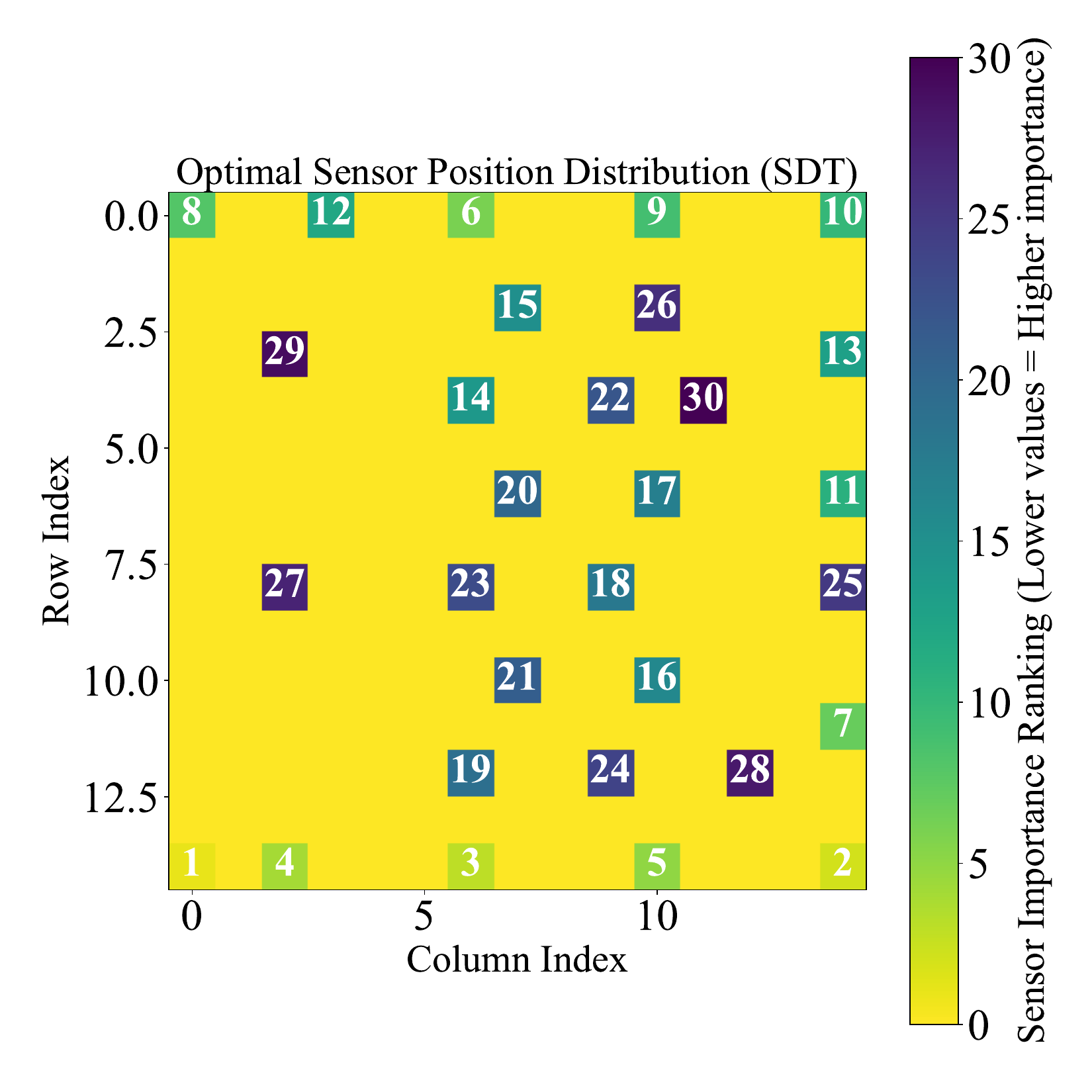}
    \hfill
    \includegraphics[width=0.48\textwidth]{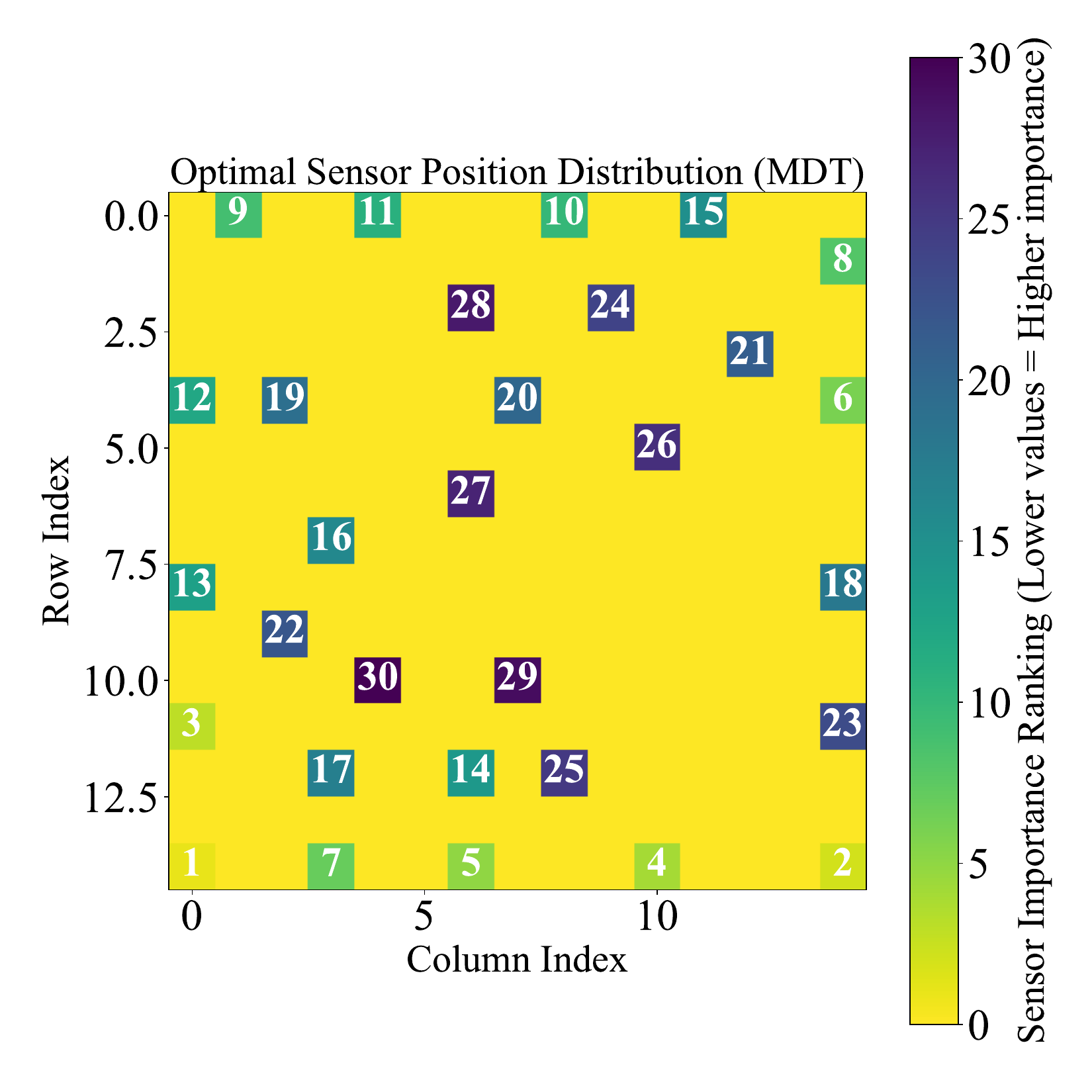}
    \caption{Sensor importance ranking results from QR analysis for (left) SDT and (right) MDT. Each panel shows the spatial distribution of the top 30 most important sensor positions, with ranking order indicated by numbers.}
    \label{fig:qr_ranking}
\end{figure}

The QR decomposition results for our dataset are presented in Fig. \ref{fig:qr_ranking} according to the two data splitting strategies, which illustrate the spatial distribution and importance ranking of sensor positions across the rooftop domain. The ranking algorithm processes all 450 potential measurement locations (225 spatial positions $\times$ 2 velocity components) and assigns importance scores based on their contribution to the POD basis linear independence.

\subsection{Optimized Sensor Performance}

The QR decomposition method was applied to identify optimal sensor positions based on POD basis functions derived from the wind field reconstruction problem. This section evaluates the effectiveness of QR-optimized sensor placement with particular emphasis on system robustness and stability under realistic operational conditions.

\subsubsection{Enhanced System Robustness}

QR-based optimization substantially enhances system robustness to sensor position uncertainties. Table~\ref{tab:qr_robustness} presents average robustness improvements across all sensor configurations for both data splitting strategies. MDT achieves higher effectiveness (90.0\% positive improvements) than SDT (60.0\% positive improvements). MDT outperforms SDT in average overall improvement (3.7\% vs. 3.1\%), demonstrating that diverse training data enhances optimal sensor placement benefits.

The robustness improvements exhibit distinct patterns across individual reconstruction methods and training strategies. SDT produces highest improvements in CWGAN (+6.5\% overall) through NMSE enhancement (+27.8\%), while Kriging achieves balanced gains (+4.1\% overall). MDT establishes a different performance hierarchy with Kriging leading (+7.9\% overall) across all metrics, followed by ViTAE (+4.8\%) with notable MG improvements (+7.7\% vs. Kriging). UNet exhibits opposite trends between strategies: SDT shows negative improvement (-0.7\%) while MDT produces positive results (+0.4\%). CWGAN demonstrates reduced effectiveness under MDT (+1.8\%) compared to SDT (+6.5\%), reflecting differential compatibility between adversarial training approaches and multi-directional flow optimization under QR constraints.

\begin{table}[htbp]
    \centering
    \caption{Average QR robustness improvements (\%) for both methods. QR improvement represents the difference in robustness (percentage performance retention under sensor perturbations) between optimized and standard placements, averaged across sensor counts (5--30).}
    \label{tab:qr_robustness}
    \setlength{\tabcolsep}{3pt}
    \begin{tabular}{lcccccccccc}
        \toprule
        \multirow{2}{*}{Strategy} & \multicolumn{5}{c}{SDT} & \multicolumn{5}{c}{MDT}                                                                \\
        \cmidrule(lr){2-6} \cmidrule(lr){7-11}
                                  & SSIM                    & NMSE                    & MG   & FAC2 & Overall & SSIM  & NMSE & MG   & FAC2 & Overall \\
        \midrule
        Kriging                   & -0.0                    & +18.1                   & -2.2 & +0.8 & +4.1    & +12.5 & +9.9 & +5.6 & +3.7 & +7.9    \\
        UNet                      & +14.2                   & -11.6                   & -6.9 & +1.4 & -0.7    & +5.1  & -1.7 & -4.7 & +3.1 & +0.4    \\
        ViTAE                     & -2.7                    & +12.2                   & -0.1 & +1.1 & +2.6    & +1.1  & +9.9 & +7.7 & +0.4 & +4.8    \\
        CWGAN                     & +0.2                    & +27.8                   & -2.1 & +0.4 & +6.5    & +1.3  & +3.2 & +1.1 & +1.5 & +1.8    \\
        \bottomrule
    \end{tabular}
\end{table}

\subsubsection{Optimization Characteristics and Performance Trade-offs}

\begin{figure*}[ht]
    \centering
    \begin{subfigure}{\textwidth}
        \centering
        \includegraphics[width=\textwidth]{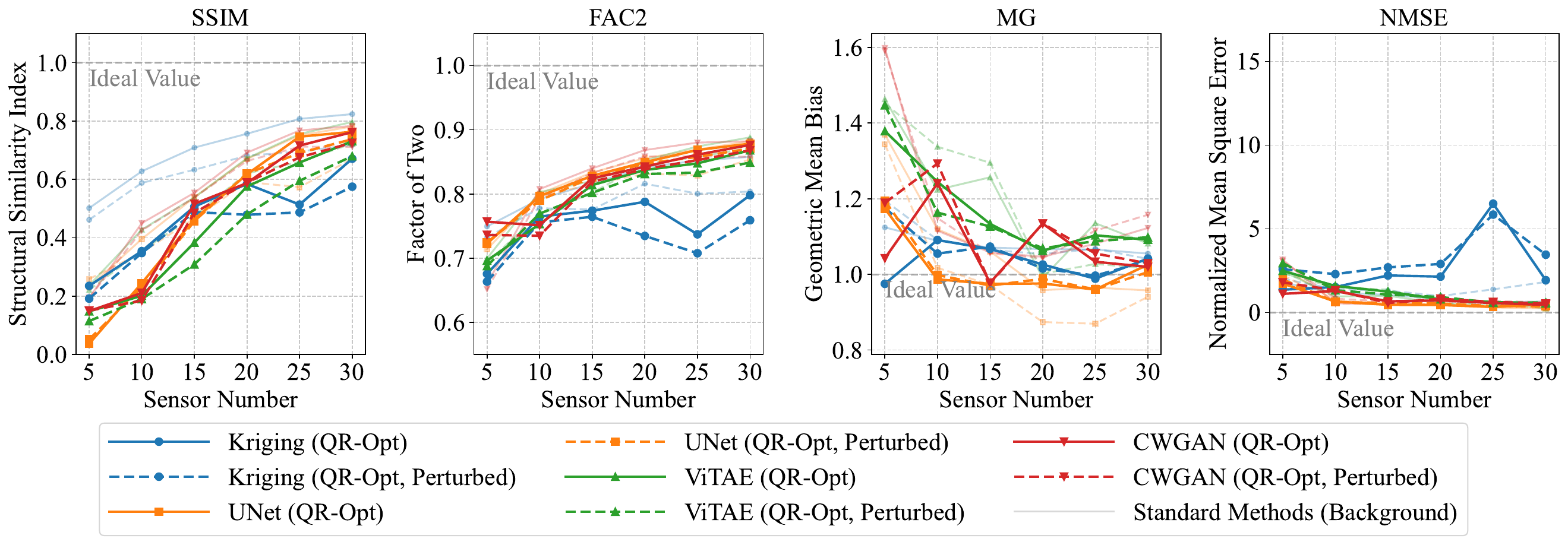}
        \caption{Comparison of Metrics Across Models with QR-based Sensor Placement Under SDT.}
        \label{fig:opt_vs_standard:a}
    \end{subfigure}

    \vspace{1em}

    \begin{subfigure}{\textwidth}
        \centering
        \includegraphics[width=\textwidth]{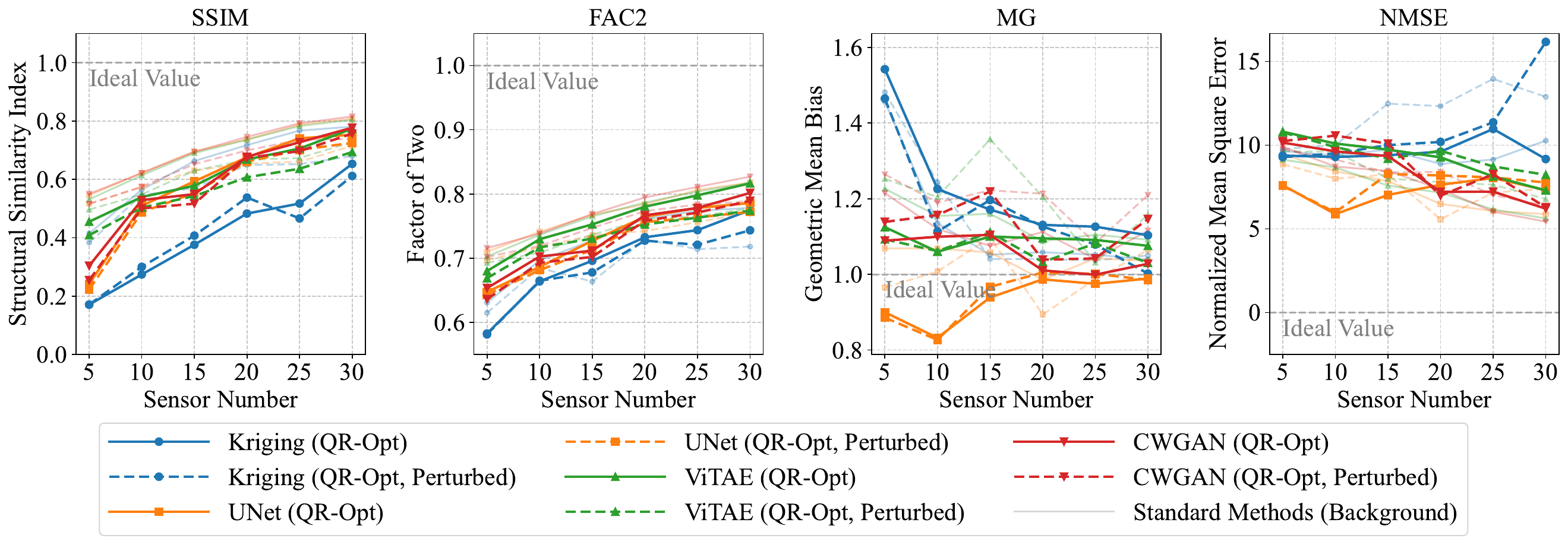}
        \caption{Comparison of Metrics Across Models with QR-based Sensor Placement Under MDT.}
        \label{fig:opt_vs_standard:b}
    \end{subfigure}

    \caption{Comparison of Metrics Across Models with QR-based Sensor Placement under different training strategies.}
    \label{fig:opt_vs_standard}
\end{figure*}

QR-based optimization demonstrates clear metric-dependent trade-offs across both training strategies in Figure~\ref{fig:opt_vs_standard}. While QR optimization enhances system robustness, it generally produces modest decreases in individual metrics. SSIM performance degrades consistently across both methods, with SDT showing decreases of 23-84\% at five sensors and MDT showing decreases of 14-59\%. SDT demonstrates superior MG improvements, with all approaches moving closer to the ideal value of 1.0 (CWGAN: 1.598→1.042, UNet: 1.370→1.174). MDT shows less pronounced MG improvements, though UNet achieves better geometric accuracy at higher sensor densities (30 sensors: 1.036→0.989). For NMSE, deep learning methods maintain comparable performance to pre-optimization levels, while Kriging suffers substantial degradation under SDT (30 sensors: 0.662→1.935, +192\% increase). FAC2 results show Kriging experiences notable decreases (7-10\% under SDT), while deep learning methods exhibit minimal degradation (0-2\%), with performance gaps diminishing at higher sensor densities where post-optimization performance approaches pre-optimization levels.

\section{Conclusions and Future Work}

Rooftop spaces accommodate diverse functions; however, their utilization is often constrained by the complex spatiotemporal variability of wind fields. Consequently, the provision of real-time wind information is essential to support more effective and reliable rooftop applications. This study established a learning-from-observation framework for real-time rooftop wind field reconstruction using PIV measurements and sparse sensors. A comprehensive benchmark was performed between Kriging interpolation and three deep learning approaches—UNet, ViTAE, and CWGAN—under two training strategies (single-direction and multi-direction). Robustness was assessed via sensor position perturbations and QR decomposition-based sensor placement.

Under the single-direction training strategy, Kriging interpolation generally surpassed deep learning methods in structural similarity and reliability. At very low sensor counts, deep learning lagged significantly behind Kriging with SSIM performance gaps of 52.7–61.4\%. This was caused by two factors: (i) the lack of diversity in the training set limited cross-direction generalization; and (ii) extremely sparse observations reduced the networks' capacity to learn informative spatial features. As the number of sensors increased, deep learning performance improved steadily: overall errors decreased and reliability increased (FAC2 generally exceeded 0.8 at higher sensor densities), narrowing the SSIM gap and even surpassing Kriging in FAC2 and overall error. In contrast, under the multi-direction training strategy, deep learning consistently outperformed Kriging across most metrics, achieving SSIM improvements of 20.0–32.7\%, FAC2 improvements of 3.5–24.2\%, and NMSE improvements of 10.2–27.8\%. Exposure to diverse wind directions during training enabled the models to learn cross-direction flow patterns, therefore outperformed Kriging over all metrics. The results indicate the necessity of multi-wind-direction training for deep learning methods to achieve better performance. \firstrevision{Further analysis in \ref{zero bad} revealed that performance variations across wind directions are partially attributable to inherent spatial characteristics of the flow field. The 0° wind direction exhibited higher boundary–center differences, spatial gradients, and sample imbalance, leading to reconstruction challenges for both Kriging and deep learning methods. This finding explains why Kriging showed a performance drop from single-direction to multi-direction training, as the multi-direction training included challenging 0° samples that were absent in single-direction evaluation.}

In the multi-directional condition, which is more realistic, deep learning methods are generally recommended for improved reliability and error reduction. UNet provides balanced accuracy with stable geometric bias (MG), serving as a reliable default for most scenarios. CWGAN achieves the highest structural similarity with robustness benefits in challenging cases, making it appropriate when high accuracy is paramount despite its notably higher computational cost. ViTAE maintains favorable efficiency with competitive accuracy, fitting edge or real-time scenarios with constrained resources. \thirdrevision{When training data are limited to a single direction and sensors are few, Kriging interpolation remains preferable, especially when structural fidelity is prioritized.} Robustness analysis showed that deep learning methods generally outperformed Kriging; however, sensitivity to perturbations tended to increase as the number of sensors increased. \thirdrevision{When robustness to sensor placement uncertainties is the primary concern (e.g., installation tolerances or expected positional drift), QR-based placement is the preferred choice; conversely, when sensor locations are fixed and precisely controlled, a uniform distribution is a better option.} These findings indicate that sensor optimization and training strategy should be considered jointly when aiming for reliable deployment. \firstrevision{Regarding temporal averaging strategies, post-averaging generally achieves better performance than pre-averaging, though the differences are relatively small and exceptions exist. Therefore, when temporal data availability is limited or computational efficiency is prioritized, pre-averaging serves as a viable alternative.}

\thirdrevision{The evaluation under realistic constraints including sensor position perturbations and sparse configurations provides confidence for practical deployment. However, several limitations remain.} \thirdrevision{The POD-QR sensor optimization method is data-driven and optimized for the specific building geometry and wind conditions in the current dataset. Generalization to different building geometries or wind regimes would require recomputing the POD basis and QR decomposition for the new scenarios. Furthermore, while incorporating multiple wind directions in the training data substantially improves model generalization across the tested directions (0°, 22.5°, and 45°), generalization beyond these specific angles remains limited, and extension to additional wind directions would require additional experimental data or alternative approaches such as transfer learning.} \fourthrevision{Additionally, the experiment considers a single isolated rectangular building at one measurement height ($z/H = 1.05$). Extension to other building geometries or measurement heights would require additional experimental data and retraining of the models.}

Future work will consider \thirdrevision{validation with real-world field measurements under natural atmospheric conditions to assess model performance beyond controlled wind tunnel environments,} next-step and multi-step temporal prediction, extension to multi-height or three-dimensional reconstruction, physics-informed objectives and constraints, and online/adaptive learning in realistic deployments.

\newpage
\section*{Acknowledgements}
Sibo Cheng acknowledges the support of the French Agence Nationale de la Recherche (ANR) under reference ANR-22-CPJ2-0143-01. CEREA is a member of Institut Pierre-Simon Laplace (IPSL). Part of this work was supported by the Japan Society for the Promotion of Science (JSPS) KAKENHI Grant Number 24K17398 (Representative: Chao Lin).

\section*{Competing interests}
The contact author has declared that none of the authors has any competing interests.

\section*{Author contributions}
Yihang Zhou: Methodology, Software, Validation, Formal analysis, Investigation, Writing - Original Draft, Writing - Review and Editing, Visualization
\vskip\baselineskip
Chao Lin: Conceptualization, Methodology, Formal analysis, Investigation, Writing - Original Draft, Writing - Review and Editing, Visualization, Funding acquisition
\vskip\baselineskip
Hideki Kikumoto: Resources, Supervision, Formal analysis, Writing - Review and Editing
\vskip\baselineskip
Ryozo Ooka: Resources, Supervision, Writing - Review and Editing
\vskip\baselineskip
Sibo Cheng: Project administration, Methodology, Software, Resources, Supervision, Formal analysis, Writing - Review and Editing

\section*{Data availability}
The code and trained models of the present paper is available at \url{https://github.com/Yng314/windreconstruction}
\appendix
\section{Detailed Analysis on 0° Wind Direction}
\setcounter{figure}{0}
\setcounter{table}{0}
\label{zero bad}
\firstrevision{Fig. \ref{fig:spatial_comparison_SDT_5sensors} and Fig. \ref{fig:spatial_comparison_MDT_5sensors} show a pattern that in 22.5° and 45° wind directions, the MDT performs better than SDT, which corresponds to our analysis in Section 4.1 and Section 4.2. However, within the MDT, the 0° wind direction shows much worse performance than 22.5° and 45°, which is unexpected and requires further investigation.}

\begin{figure*}[ht]
    \centering
    \includegraphics[width=\textwidth]{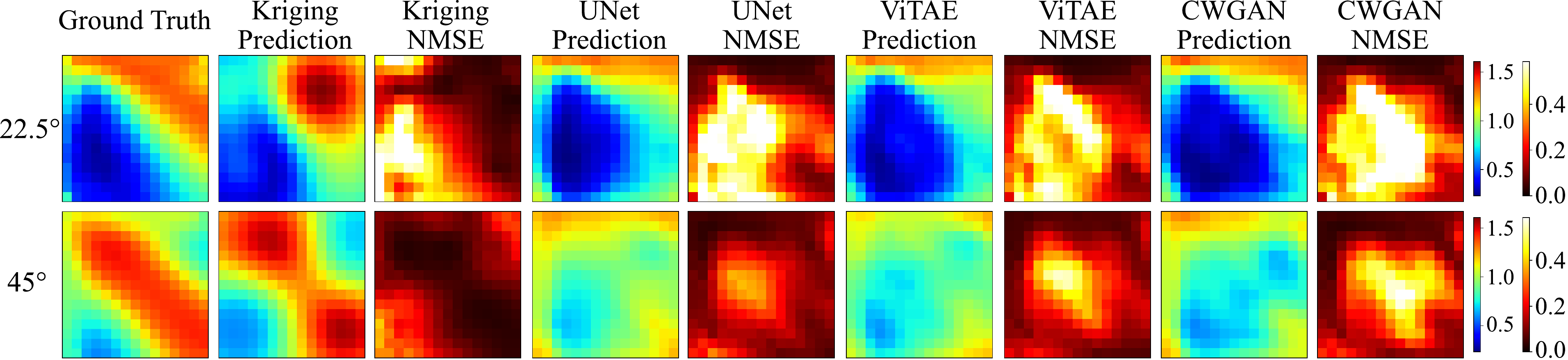}
    \captionsetup{labelformat=empty}
    \caption{\firstrevision{Figure A1: Spatial NMSE distribution of 0° wind direction for SDT with 5 sensors helps visualize which part of the wind field is difficult to predict. With low sensor density, Kriging exhibits better performance compared to deep learning methods, which corresponds to Fig. \ref{fig:standard:a}.}}
    \label{fig:spatial_comparison_SDT_5sensors}
\end{figure*}

\begin{figure*}[ht]
    \centering
    \includegraphics[width=\textwidth]{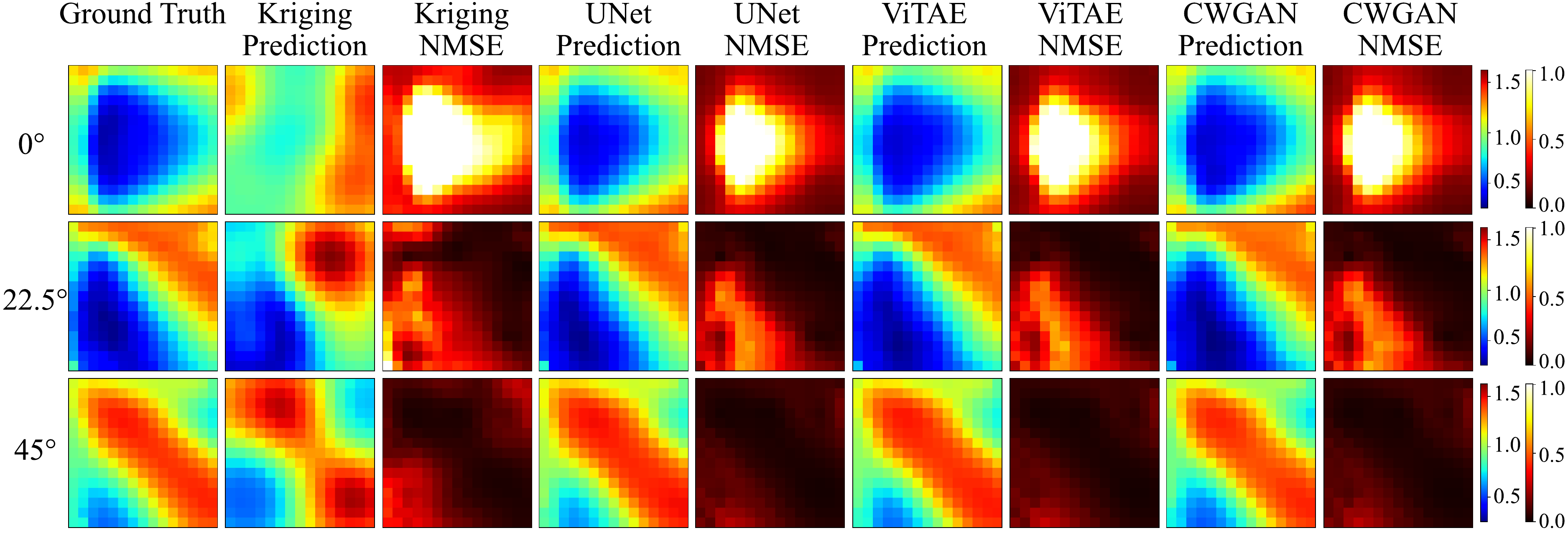}
    \caption{\firstrevision{Spatial NMSE distribution of 0° wind direction for MDT with 5 sensors. Even with low sensor density, deep learning methods still show potential advantage over Kriging, especially for 0° wind direction, which corresponds to Fig. \ref{fig:standard:b}.}}
    \label{fig:spatial_comparison_MDT_5sensors}
\end{figure*}

\begin{figure*}[ht]
    \centering
    \includegraphics[width=\textwidth]{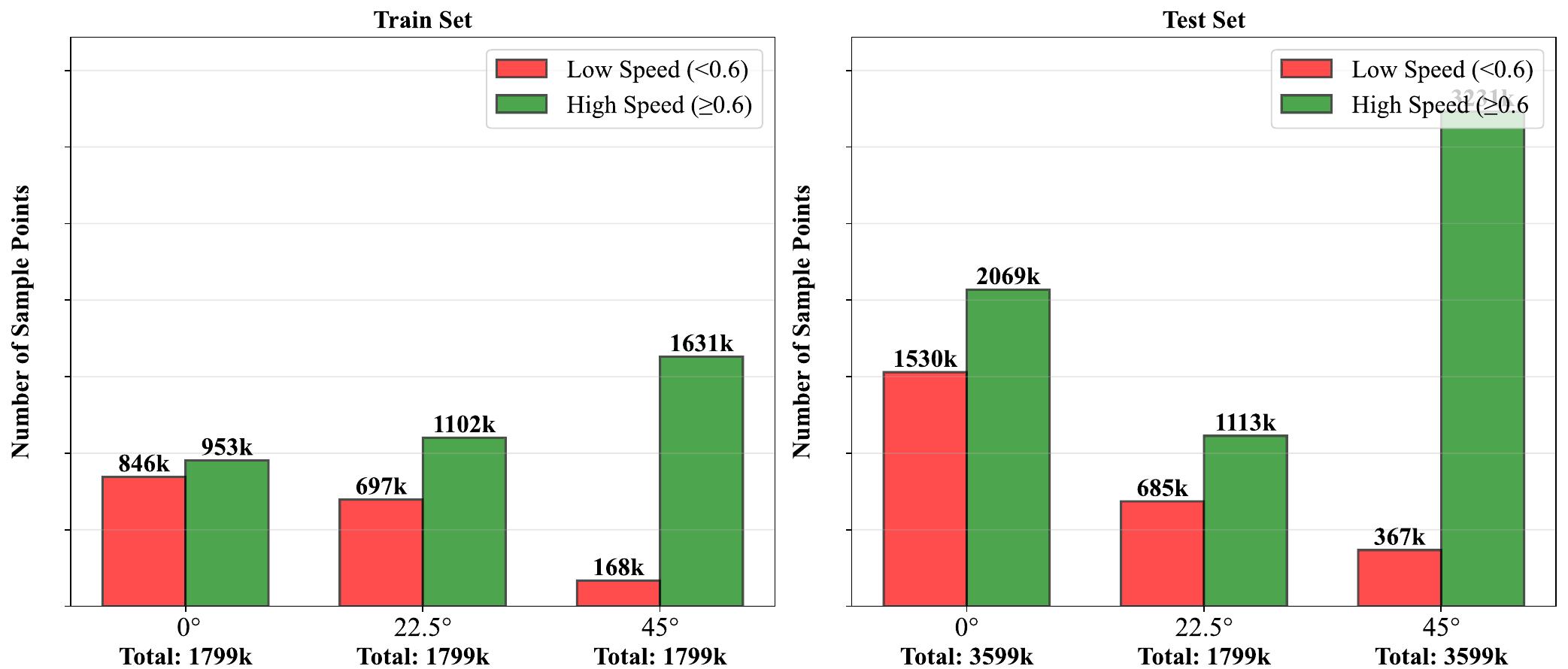}
    \caption{\firstrevision{High-speed and low-speed sample counts in the training and testing datasets using a threshold of 0.6 m/s. All spatial points in each wind field snapshot are counted, with velocities greater than or equal to 0.6 m/s classified as high-speed and velocities below 0.6 m/s classified as low-speed.}}
    \label{fig:train_test_sample_counts_threshold_0_6}
\end{figure*}

\firstrevision{After analyzing the three different wind direction datasets, we propose two possible reasons for the poor performance of 0° wind direction using either Kriging or deep learning methods. Firstly, we calculate the boundary–center difference, coefficient of variation (CV) and spatial gradient of the three wind direction datasets, and the results are shown in Table \ref{tab:spatial_features}. We can see that 0° wind direction has the largest boundary–center difference, which indicates strong non-stationarity, a potential reason for the poor performance of 0° wind direction using Kriging. Besides, the largest spatial gradient and high CV of 0° wind direction make it difficult to predict using deep learning methods. These metrics indicate that the poor performance of 0° wind direction was partially caused by the distinct spatial features of this dataset. Secondly, in Fig. \ref{fig:spatial_comparison_MDT_5sensors}, the NMSE distributions show a pattern that high NMSE values are concentrated in the blue region of the ground truth (around 0.6), indicating the low-speed region is difficult to predict. After taking 0.6 as the threshold and plotting the wind speed sample distribution of the training and testing datasets in Fig. \ref{fig:train_test_sample_counts_threshold_0_6}, we can see that the training samples for high speed are roughly twice the size of those for low speed, which may cause an imbalance issue, leading to the poor deep learning performance in the low-speed region. Given this imbalance issue and that the 0° wind direction contains most of the low-speed samples in the testing dataset, it is not surprising that the deep learning methods perform poorly at 0° wind direction.}

\firstrevision{In summary, the poor performance of 0° wind direction using deep learning can be attributed to the dataset characteristics and the potential imbalance issue. As for Kriging, the strong non-stationarity of the 0° wind direction dataset leads to the poor performance for this wind direction. This finding also explains why Kriging showed a performance drop from SDT to MDT, since Kriging performs poorly at interpolating 0° wind direction while there are no 0° data to be interpolated in SDT (SDT used all 0° data for the training set).}

\begin{table}[ht]
    \centering
    \caption{\firstrevision{Spatial features of 0°, 22.5° and 45° wind directions. \secondrevision{Boundary–center difference measures spatial non-stationarity; high values challenge Kriging's stationarity assumption. Coefficient of variation quantifies velocity heterogeneity; high CV indicates coexisting high and low speed regions that complicate reconstruction. Spatial gradient measures the rate of velocity change; high gradients create sharp transitions difficult to capture from sparse sensors.}}}
    \begin{tabular}{llll}
        \hline
        Feature                       & 0°    & 22.5° & 45°   \\ \hline
        Boundary–center difference    & 0.772 & 0.210 & 0.372 \\
        Coefficient of variation (CV) & 0.507 & 0.566 & 0.229 \\
        Spatial gradient              & 0.143 & 0.127 & 0.096 \\ \hline
    \end{tabular}
    \label{tab:spatial_features}
\end{table}

\section{Analysis on Asymmetric Sensor Placement in the SDT}
\setcounter{figure}{0}
\setcounter{table}{0}
\firstrevision{In the QR-optimized sensor placement of SDT, we found that the sensor placement is not symmetric compared to the PIV setup in Fig.~\ref{fig:WTE_Results}, especially the top 10 sensors visualized in Fig.~\ref{fig:optimal_sensors_method0_num10}, where the top five sensors are concentrated in the bottom of the field instead of being evenly distributed. To investigate this unexpected result, we calculated the variance of the 0° wind direction dataset and plotted it in Fig.~\ref{fig:0deg_wind_speed_variance}. The result shows that the variance is highest in the bottom of the field, which is consistent with the top 5 sensor placement. This explains why the top five sensors are concentrated in the bottom of the field, as the QR optimization process prioritizes regions with higher variance. This asymmetric sensor distribution reflects the inherent characteristics of experimental data, which exhibit spatial variability influenced by real-world conditions rather than the idealized uniformity often observed in numerical simulations. The variance-driven optimization therefore provides insights into the spatial heterogeneity of experimental wind fields and demonstrates the value of using experimental data for wind field reconstruction studies.}

\begin{figure*}[ht]
    \centering
    \subcaptionbox{\label{fig:optimal_sensors_method0_num10}}
    {\includegraphics[width=0.48\textwidth]{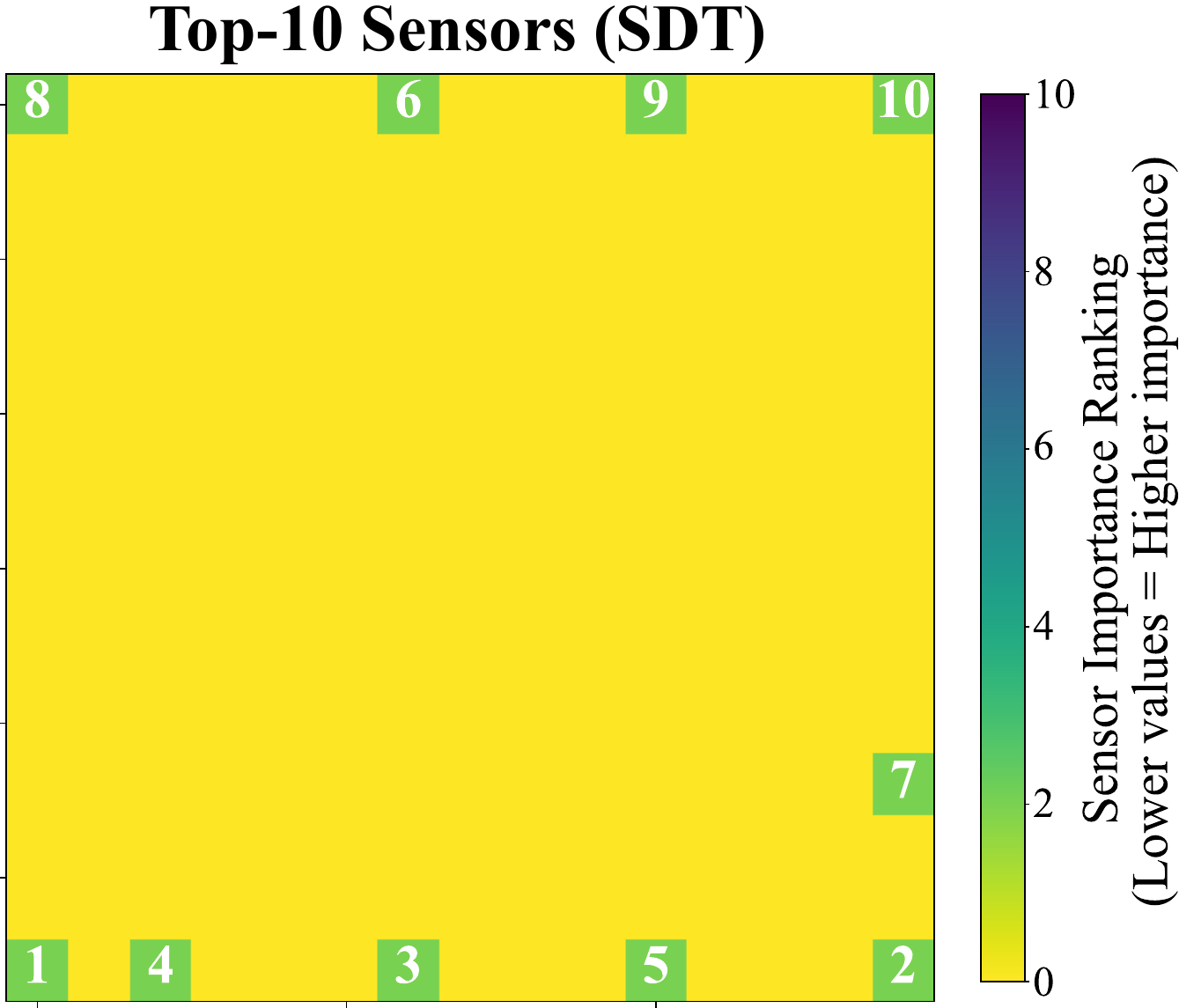}}
    \hfill
    \subcaptionbox{\label{fig:0deg_wind_speed_variance}}
    {\includegraphics[width=0.48\textwidth]{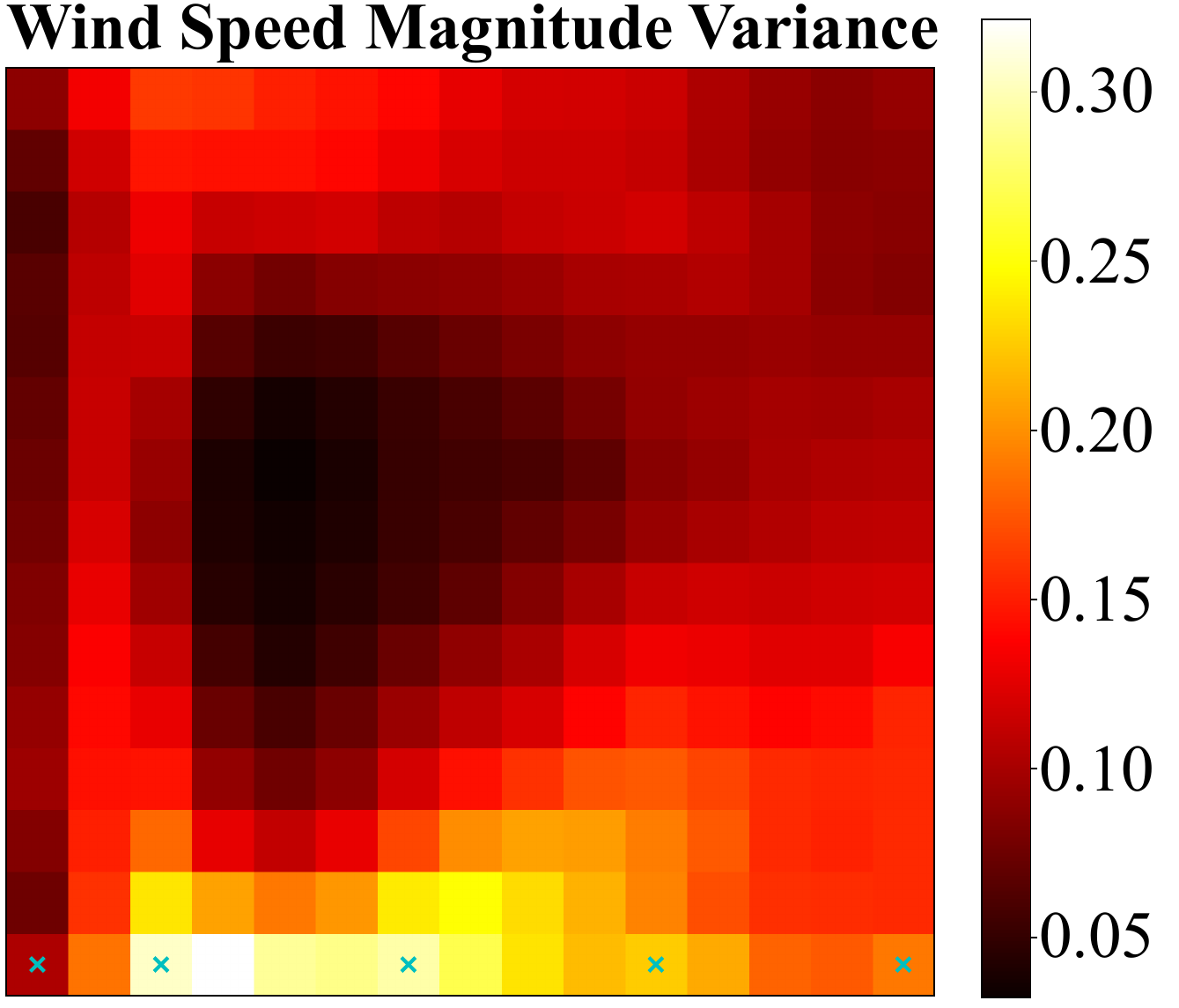}}
    \caption{\firstrevision{Analysis of asymmetric sensor placement: (a) Top 10 sensor placement in the SDT after QR-based optimization. (b) Variance of the 0° wind direction dataset, green crosses indicate the top 5 sensors. There is a clear pattern that the top 5 sensors are concentrated in the bottom of the field, where the variance is highest.}}
    \label{fig:sensor_placement_variance}
\end{figure*}

\section{CWGAN Multiple Generation Comparison}
\setcounter{figure}{0}
\setcounter{table}{0}
One of the features of CWGAN is its potential to generate diverse samples. We averaged the predictions of CWGAN over different noise realizations to evaluate this variability. The results under SDT with QR optimized sensor placement, shown in Table \ref{tab:cwgan_averaging_comparison_standard} and Fig.~\ref{fig:cwgan_averaging_comparison_standard}, indicate that the outputs are very close, with minimal distribution differences. \thirdrevision{This limited variability is primarily attributed to the strong weight assigned to the L1 loss term in our objective function. Our experiments indicated that a strong L1 constraint is necessary to achieve the high point-wise reconstruction accuracy required for safety-critical wind field applications, though this comes at the cost of reduced generative diversity. To further investigate this trade-off, we conducted an ablation study comparing the proposed model (High L1) with a variant using a reduced L1 weight (Low L1). The results in Table \ref{tab:l1_comparison} show that while reducing the L1 weight might increase variability to some extent, it leads to significant performance degradation. Therefore, in this specific configuration, the adversarial component functions primarily as a regularizer to enhance structural fidelity (e.g., higher SSIM) rather than as a driver for stochastic uncertainty, representing a trade-off between reconstruction precision and generative variability.}

\begin{table}[htbp]
    \centering
    \caption{\thirdrevision{Performance comparison of CWGAN with reduced L1 weight (Low L1) versus the proposed configuration (High L1). The results show that relaxing the L1 constraint leads to performance degradation.}}
    \label{tab:l1_comparison}
    \begin{tabular}{llcccc}
        \toprule
        Sensor Counts & Metrics & \multicolumn{2}{c}{Original (Single Run)} & \multicolumn{2}{c}{30\_avg (Ensemble)}                               \\
        \cmidrule(lr){3-4} \cmidrule(lr){5-6}
                      &         & Low L1                                    & High L1 (Proposed)                     & Low L1 & High L1 (Proposed) \\
        \midrule
        5             & MG      & 1.4357                                    & \textbf{1.0423}                        & 1.4420 & \textbf{1.0421}    \\
                      & NMSE    & 2.9615                                    & \textbf{1.1196}                        & 3.0701 & \textbf{1.1553}    \\
                      & FAC2    & 0.6704                                    & \textbf{0.7568}                        & 0.6706 & \textbf{0.7568}    \\
                      & SSIM    & 0.1406                                    & \textbf{0.1487}                        & 0.1413 & \textbf{0.1487}    \\
        \midrule
        25            & MG      & 0.9914                                    & 1.0333                                 & 0.9924 & 1.0333             \\
                      & NMSE    & 0.5695                                    & \textbf{0.5671}                        & 0.5759 & 0.5802             \\
                      & FAC2    & 0.8425                                    & \textbf{0.8612}                        & 0.8439 & \textbf{0.8613}    \\
                      & SSIM    & 0.6870                                    & \textbf{0.7158}                        & 0.6917 & \textbf{0.7159}    \\
        \bottomrule
    \end{tabular}
\end{table}

\begin{table}[htbp]
    \centering
    \caption{\firstrevision{Comparison of CWGAN performance averaged over different prediction samples. Bold values indicate the best performance, though the differences between them are minimal.}}
    \begin{tabular}{llllll}
        \hline
        Sensor Counts & Metrics & Original        & 5\_avg          & 10\_avg         & 30\_avg         \\ \hline
        5             & MG      & 1.0423          & 1.0427          & 1.0422          & \textbf{1.0421} \\
                      & NMSE    & \textbf{1.1196} & 1.1427          & 1.1466          & 1.1553          \\
                      & FAC2    & 0.7568          & 0.7568          & 0.7568          & 0.7568          \\
                      & SSIM    & 0.1487          & 0.1487          & 0.1487          & 0.1487          \\
        10            & MG      & 1.2406          & \textbf{1.2349} & 1.2399          & 1.2446          \\
                      & NMSE    & 1.2898          & 1.2482          & 1.2481          & \textbf{1.245}  \\
                      & FAC2    & 0.7512          & 0.7512          & 0.7512          & 0.7512          \\
                      & SSIM    & 0.2106          & 0.2106          & 0.2106          & 0.2106          \\
        15            & MG      & \textbf{0.9779} & \textbf{0.9779} & 0.977           & 0.9758          \\
                      & NMSE    & 0.6714          & \textbf{0.6243} & 0.6267          & 0.6473          \\
                      & FAC2    & 0.8244          & \textbf{0.8248} & 0.8247          & 0.8247          \\
                      & SSIM    & 0.5163          & \textbf{0.5176} & \textbf{0.5176} & \textbf{0.5176} \\
        20            & MG      & 1.1342          & 1.1359          & \textbf{1.1336} & 1.1347          \\
                      & NMSE    & 0.7278          & \textbf{0.7156} & 0.7221          & 0.7194          \\
                      & FAC2    & 0.8425          & 0.8425          & 0.8425          & 0.8425          \\
                      & SSIM    & 0.5873          & 0.5873          & 0.5873          & 0.5873          \\
        25            & MG      & \textbf{1.0333} & 1.0337          & 1.0335          & \textbf{1.0333} \\
                      & NMSE    & \textbf{0.5671} & 0.5777          & 0.5825          & 0.5802          \\
                      & FAC2    & 0.8612          & \textbf{0.8613} & \textbf{0.8613} & \textbf{0.8613} \\
                      & SSIM    & 0.7158          & \textbf{0.7159} & \textbf{0.7159} & \textbf{0.7159} \\
        30            & MG      & \textbf{1.0197} & 1.0199          & 1.02            & 1.02            \\
                      & NMSE    & 0.464           & \textbf{0.461}  & 0.4611          & 0.4645          \\
                      & FAC2    & 0.8764          & 0.8764          & 0.8764          & 0.8764          \\
                      & SSIM    & 0.7621          & 0.7621          & 0.7621          & 0.7621          \\ \hline
    \end{tabular}

    \label{tab:cwgan_averaging_comparison_standard}
\end{table}

\begin{figure*}[ht]
    \centering
    \includegraphics[width=\textwidth]{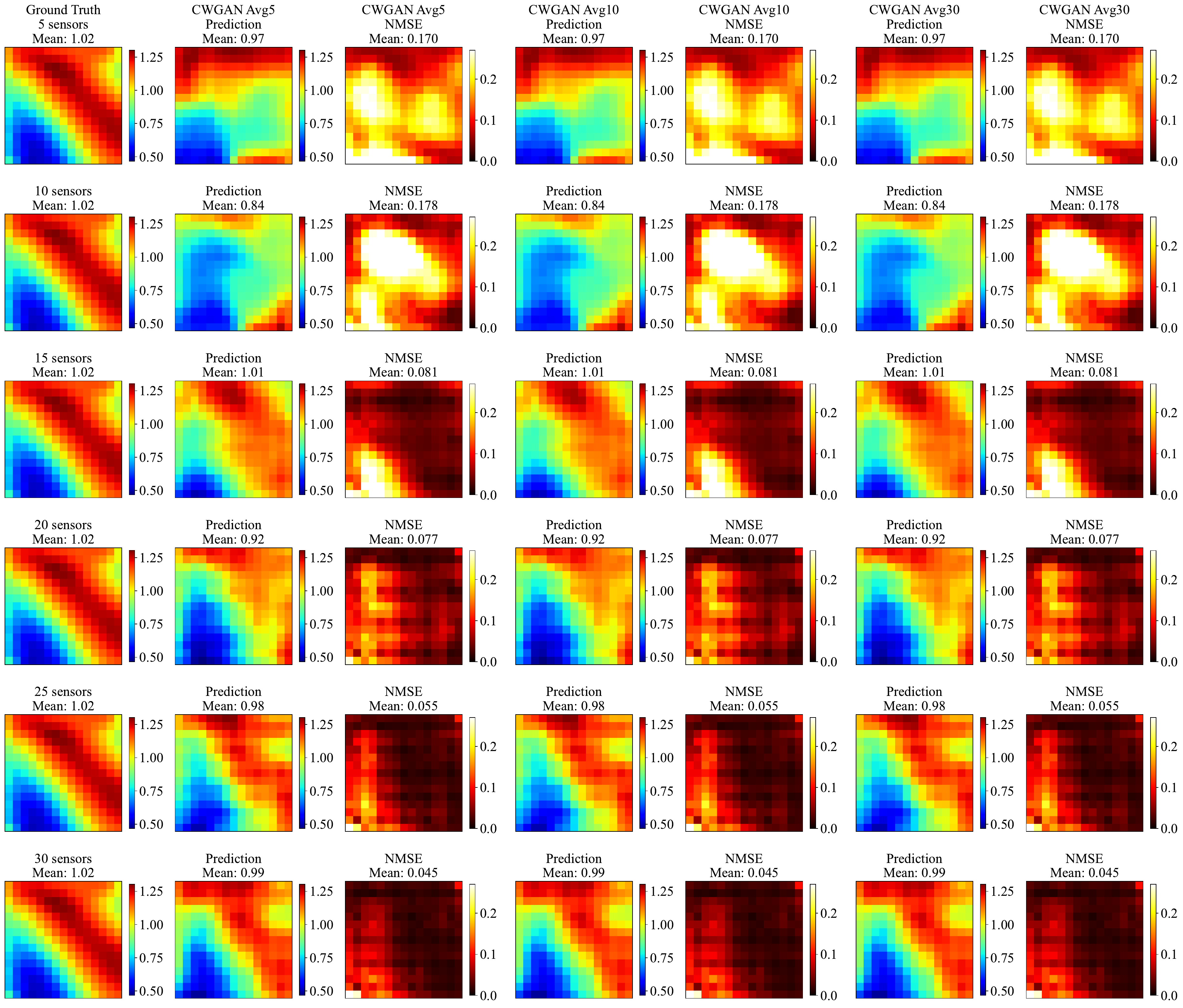}
    \caption{\firstrevision{Comparison of CWGAN performance averaged over different prediction samples. The differences are visually negligible.}}
    \label{fig:cwgan_averaging_comparison_standard}
\end{figure*}



\section{Robustness Analysis of MDT Strategy}
\label{appendix:mdt_robustness}
\setcounter{figure}{0}
\setcounter{table}{0}

\subsection{Similarity Analysis Between Realizations}
\label{appendix:similarity}

\thirdrevision{Our similarity analysis among different realizations under the same wind direction using SSIM to validate the data splitting strategy shown in Table \ref{tab:realization_similarity} demonstrates that realizations within the same wind direction exhibit high structural similarity (average SSIM: 0.9677 for 0°, 0.9835 for 22.5°, 0.9588 for 45°), confirming that they capture consistent flow patterns characteristic of each wind direction. However, the SSIM values remain below 1.0 (with 3-4\% structural differences), demonstrating that these realizations are independent experimental sessions rather than identical duplicates. This validates our data splitting strategy, where one realization is used for training and others for testing, ensuring genuine model evaluation while maintaining physical consistency within each wind direction.}

\begin{table}[htbp]
    \centering
    \caption{\thirdrevision{Structural similarity (SSIM) between different realizations ($\mathcal{D}_{\theta}^{(1)}, \mathcal{D}_{\theta}^{(2)}, \mathcal{D}_{\theta}^{(3)}$) within the same wind direction ($\theta$). Values close to 1.0 indicate high similarity, while values below 1.0 (with 3-4\% structural differences) confirm independence between realizations.}}
    \label{tab:realization_similarity}
    \begin{tabular}{lcccc}
        \toprule
        \diagbox[width=4cm]{Direction $\theta$}{Realization} & $\mathcal{D}_{\theta}^{(1)}$ vs $\mathcal{D}_{\theta}^{(2)}$ & $\mathcal{D}_{\theta}^{(1)}$ vs $\mathcal{D}_{\theta}^{(3)}$ & $\mathcal{D}_{\theta}^{(2)}$ vs $\mathcal{D}_{\theta}^{(3)}$ & Average SSIM \\
        \midrule
        0°                                                   & 0.9605                                                       & 0.9620                                                       & 0.9807                                                       & 0.9677       \\
        22.5°                                                & 0.9835                                                       & --                                                           & --                                                           & 0.9835       \\
        45°                                                  & 0.9523                                                       & 0.9480                                                       & 0.9761                                                       & 0.9588       \\
        \bottomrule
    \end{tabular}
\end{table}

\subsection{Sensitivity to Training Realization Selection}
\label{appendix:sensitivity}

\thirdrevision{We conducted a sensitivity analysis by changing which run is used for training under MDT, the results are shown in Table \ref{tab:mdt_sensitivity}. Specifically, we compared the original split (Realization 1 for training) with an alternative split (Switch realization 2 for training). Regarding NMSE, as a variance-normalized metric whose denominator depends on the mean values of the test set, it is inherently sensitive to changes in test set composition and therefore not directly comparable across different splits. Our calculation shows that the lower variance in the new test set introduces a theoretical baseline increase of approximately 9.21\% in NMSE even with identical absolute prediction errors. For MG, most variations remain within $\pm$7\%, which is acceptable given MG's sensitivity to outliers. Most importantly, SSIM and FAC2 demonstrate high stability with variations mostly within $\pm$2\%, confirming that model performance does not depend on the particular choice of training realization and validating the robustness of our conclusions.}

\begin{table}[htbp]
    \centering
    \caption{\thirdrevision{Sensitivity analysis of MDT split by changing training realization selection. Values represent percentage differences between the alternative split (Realization 2 for training) and the original split (Realization 1 for training). SSIM and FAC2 show minimal variation (mostly within $\pm$2\%), while NMSE variations include a theoretical baseline offset of approximately 9.21\% due to test set composition changes.}}
    \label{tab:mdt_sensitivity}
    \resizebox{\textwidth}{!}{%
        \begin{tabular}{clllll}
            \toprule
            Sensor              & Model   & SSIM Diff\% & FAC2 Diff\% & NMSE Diff\% & MG Diff\% \\
            \midrule
            \multirow{4}{*}{5}  & UNet    & $-4.24$\%   & $-1.01$\%   & $7.46$\%    & $-2.48$\% \\
                                & CWGAN   & $6.25$\%    & $1.89$\%    & $-6.46$\%   & $-5.90$\% \\
                                & ViTAE   & $-1.36$\%   & $0.12$\%    & $13.49$\%   & $-9.72$\% \\
                                & Kriging & $-5.36$\%   & $-0.30$\%   & $11.69$\%   & $-2.06$\% \\
            \midrule
            \multirow{4}{*}{10} & UNet    & $-2.39$\%   & $-0.51$\%   & $15.61$\%   & $-1.26$\% \\
                                & CWGAN   & $-1.28$\%   & $0.60$\%    & $14.38$\%   & $-0.09$\% \\
                                & ViTAE   & $-0.37$\%   & $0.06$\%    & $4.00$\%    & $-3.81$\% \\
                                & Kriging & $-0.86$\%   & $0.22$\%    & $7.62$\%    & $0.32$\%  \\
            \midrule
            \multirow{4}{*}{15} & UNet    & $-2.15$\%   & $-0.53$\%   & $7.80$\%    & $-1.56$\% \\
                                & CWGAN   & $4.51$\%    & $2.18$\%    & $-9.89$\%   & $-1.47$\% \\
                                & ViTAE   & $-1.13$\%   & $-0.35$\%   & $6.54$\%    & $-6.49$\% \\
                                & Kriging & $-0.81$\%   & $0.21$\%    & $8.18$\%    & $1.40$\%  \\
            \midrule
            \multirow{4}{*}{20} & UNet    & $-1.81$\%   & $0.19$\%    & $20.12$\%   & $5.62$\%  \\
                                & CWGAN   & $-1.93$\%   & $-0.30$\%   & $5.89$\%    & $5.87$\%  \\
                                & ViTAE   & $-0.67$\%   & $0.18$\%    & $3.12$\%    & $-0.99$\% \\
                                & Kriging & $-0.56$\%   & $0.26$\%    & $6.21$\%    & $1.65$\%  \\
            \midrule
            \multirow{4}{*}{25} & UNet    & $-1.62$\%   & $-0.61$\%   & $21.36$\%   & $-2.38$\% \\
                                & CWGAN   & $0.65$\%    & $0.17$\%    & $8.23$\%    & $0.78$\%  \\
                                & ViTAE   & $-0.26$\%   & $0.02$\%    & $-2.12$\%   & $-3.12$\% \\
                                & Kriging & $-0.05$\%   & $0.22$\%    & $11.87$\%   & $-0.11$\% \\
            \midrule
            \multirow{4}{*}{30} & UNet    & $-1.63$\%   & $-0.79$\%   & $0.07$\%    & $-3.51$\% \\
                                & CWGAN   & $0.14$\%    & $-0.43$\%   & $5.02$\%    & $-6.88$\% \\
                                & ViTAE   & $-0.51$\%   & $-0.31$\%   & $13.87$\%   & $-2.72$\% \\
                                & Kriging & $-0.32$\%   & $0.04$\%    & $8.06$\%    & $1.40$\%  \\
            \bottomrule
        \end{tabular}%
    }
\end{table}

\subsection{Wind Direction Interpolation Capability}
\label{appendix:interpolation}

\thirdrevision{We conducted an additional experiment training only on 0° and 45° (excluding 22.5°) while maintaining the same test set to assess the wind direction interpolation capability of different methods, the results are shown in Table \ref{tab:exclude22}. As expected, Kriging performance remains unchanged (0\% difference) since it does not rely on training data. For deep learning methods, excluding 22.5° from training leads to modest performance degradation: SSIM decreases by 5-10\% at sparse sensor configurations (5 sensors) and 1-3\% at denser configurations (25-30 sensors), while FAC2 shows comparable trends (2-3\% degradation at sparse configurations, 1-2\% at dense configurations). Notably, at high sensor densities (25-30 sensors), some models (e.g., ViTAE) even show slight performance improvements in certain metrics, suggesting that the models can interpolate between 0° and 45° to some extent. However, the overall performance degradation confirms that including all wind directions in the training set remains the recommended approach for optimal reconstruction accuracy, particularly when sensor coverage is limited.}

\begin{table}[htbp]
    \centering
    \caption{\thirdrevision{Performance comparison between excluding 22.5° from training (0° + 45° only) versus the original MDT (0° + 22.5° + 45°). Values represent percentage differences. Kriging shows 0\% difference as it does not rely on training data. Deep learning methods show modest degradation, confirming the value of multi-direction training.}}
    \label{tab:exclude22}
    \resizebox{\textwidth}{!}{%
        \begin{tabular}{clllll}
            \toprule
            Sensor              & Model   & SSIM Diff\% & FAC2 Diff\% & NMSE Diff\% & MG Diff\% \\
            \midrule
            \multirow{4}{*}{5}  & UNet    & $-9.34$\%   & $-2.92$\%   & $3.43$\%    & $4.06$\%  \\
                                & CWGAN   & $-10.40$\%  & $-3.18$\%   & $0.83$\%    & $-6.26$\% \\
                                & ViTAE   & $-5.66$\%   & $-2.58$\%   & $1.60$\%    & $-0.19$\% \\
                                & Kriging & $0.00$\%    & $0.00$\%    & $0.00$\%    & $0.00$\%  \\
            \midrule
            \multirow{4}{*}{10} & UNet    & $-7.23$\%   & $-2.14$\%   & $5.93$\%    & $0.16$\%  \\
                                & CWGAN   & $-7.05$\%   & $-1.44$\%   & $5.45$\%    & $-2.23$\% \\
                                & ViTAE   & $-6.87$\%   & $-2.20$\%   & $1.44$\%    & $-1.53$\% \\
                                & Kriging & $0.00$\%    & $0.00$\%    & $-0.00$\%   & $0.00$\%  \\
            \midrule
            \multirow{4}{*}{15} & UNet    & $-3.58$\%   & $-1.89$\%   & $-2.10$\%   & $1.18$\%  \\
                                & CWGAN   & $-2.87$\%   & $-1.94$\%   & $-11.14$\%  & $13.66$\% \\
                                & ViTAE   & $-2.67$\%   & $-1.45$\%   & $4.56$\%    & $-1.46$\% \\
                                & Kriging & $0.00$\%    & $0.00$\%    & $-0.00$\%   & $0.00$\%  \\
            \midrule
            \multirow{4}{*}{20} & UNet    & $-2.93$\%   & $-1.56$\%   & $10.01$\%   & $5.04$\%  \\
                                & CWGAN   & $-1.61$\%   & $-1.71$\%   & $-4.22$\%   & $5.38$\%  \\
                                & ViTAE   & $-1.66$\%   & $-1.30$\%   & $-0.14$\%   & $-0.43$\% \\
                                & Kriging & $0.00$\%    & $0.00$\%    & $0.00$\%    & $0.00$\%  \\
            \midrule
            \multirow{4}{*}{25} & UNet    & $-2.15$\%   & $-1.69$\%   & $8.70$\%    & $0.86$\%  \\
                                & CWGAN   & $-1.97$\%   & $-1.46$\%   & $9.58$\%    & $5.91$\%  \\
                                & ViTAE   & $1.92$\%    & $0.41$\%    & $-9.22$\%   & $-1.61$\% \\
                                & Kriging & $0.00$\%    & $0.00$\%    & $-0.00$\%   & $0.00$\%  \\
            \midrule
            \multirow{4}{*}{30} & UNet    & $-2.80$\%   & $-1.96$\%   & $3.65$\%    & $2.16$\%  \\
                                & CWGAN   & $-0.95$\%   & $-1.47$\%   & $9.56$\%    & $-7.22$\% \\
                                & ViTAE   & $0.44$\%    & $-0.58$\%   & $-6.67$\%   & $-0.77$\% \\
                                & Kriging & $0.00$\%    & $0.00$\%    & $0.00$\%    & $-0.00$\% \\
            \bottomrule
        \end{tabular}%
    }
\end{table}

\clearpage
\clearpage
\renewcommand{\bibname}{References}
\bibliographystyle{elsarticle-num-names}
\bibliography{main.bib}

@article{doddipatla2021wind,
  title     = {Wind loads on roof-mounted equipment on low-rise buildings with low-slope roofs},
  author    = {Doddipatla, Lakshmana S and Kopp, Gregory A},
  journal   = {Journal of Wind Engineering and Industrial Aerodynamics},
  volume    = {211},
  pages     = {104552},
  year      = {2021},
  publisher = {Elsevier}
}

@article{maurer2023comparing,
  title     = {Comparing PV-green and PV-cool roofs to diverse rooftop options using decision analysis},
  author    = {Maurer, Bettina and Lienert, Judit and Cook, Lauren M},
  journal   = {Building and Environment},
  volume    = {245},
  pages     = {110922},
  year      = {2023},
  publisher = {Elsevier}
}

@article{fleck2022urban,
  title     = {Urban green roofs to manage rooftop microclimates: A case study from Sydney, Australia},
  author    = {Fleck, R and Gill, RL and Saadeh, S and Pettit, T and Wooster, E and Torpy, F and Irga, P},
  journal   = {Building and Environment},
  volume    = {209},
  pages     = {108673},
  year      = {2022},
  publisher = {Elsevier}
}

@article{lin2025wind,
  title     = {Wind Tunnel Study of Parapet Effects on Rooftop Wind Environment with Implications for Safe Urban-Air-Mobility Operations},
  author    = {Lin, Chao and Ooka, Ryozo and Takakuwa, Yasutomo and Kikumoto, Hideki},
  journal   = {Building and Environment},
  pages     = {113509},
  year      = {2025},
  publisher = {Elsevier}
}

@article{hu2022estimation,
  title     = {Estimation of airflow distribution in cubic building group model using POD-LSE and limited sensors},
  author    = {Hu, C. and Jia, H. and Kikumoto, H.},
  journal   = {Building and Environment},
  volume    = {221},
  pages     = {109324},
  year      = {2022},
  publisher = {Elsevier}
}

@article{zhang2022towards,
  title     = {Towards real-time prediction of velocity field around a building using generative adversarial networks based on the surface pressure from sparse sensor networks},
  author    = {Zhang, B. and Ooka, R. and Kikumoto, H. and Hu, C. and Tse, T. K.},
  journal   = {Journal of Wind Engineering and Industrial Aerodynamics},
  volume    = {231},
  pages     = {105243},
  year      = {2022},
  publisher = {Elsevier}
}

@article{hu2023estimation,
  title     = {Estimation of instantaneous airflow distribution in cubic building group model using multi-time-delay LSE-POD},
  author    = {Hu, C. and Jia, H. and Kikumoto, H.},
  journal   = {Building and Environment},
  volume    = {243},
  pages     = {110642},
  year      = {2023},
  publisher = {Elsevier}
}

@article{hu2024fast,
  title     = {Fast estimation of airflow distribution in an urban model using generative adversarial networks with limited sensing data},
  author    = {Hu, C. and Kikumoto, H. and Zhang, B. and Jia, H.},
  journal   = {Building and Environment},
  volume    = {249},
  pages     = {111120},
  year      = {2024},
  publisher = {Elsevier}
}

@article{watkins2020ten,
  title     = {Ten questions concerning the use of drones in urban environments},
  author    = {Watkins, Simon and Burry, Jane and Mohamed, Abdulghani and Marino, Matthew and Prudden, Samuel and Fisher, Alex and Kloet, Nicola and Jakobi, Timothy and Clothier, Reece},
  journal   = {Building and Environment},
  volume    = {167},
  pages     = {106458},
  year      = {2020},
  publisher = {Elsevier}
}

@incollection{castagno2021map,
  title     = {Map-based planning for small unmanned aircraft rooftop landing},
  author    = {Castagno, J and Atkins, E},
  booktitle = {Handbook of Reinforcement Learning and Control},
  pages     = {613--646},
  year      = {2021},
  publisher = {Springer}
}

@article{pu2023research,
  title     = {Research on the characteristics of urban building cluster wind field based on UAV wind measurement},
  author    = {Pu, Ou and Yuan, Boqiu and Li, Zhengnong and Bao, Terigen and Chen, Zheng and Yang, Liwei and Qin, Hua and Li, Zhen},
  journal   = {Buildings},
  volume    = {13},
  number    = {12},
  pages     = {3109},
  year      = {2023},
  publisher = {MDPI}
}

@article{carpentieri2015influence,
  title     = {Influence of urban morphology on air flow over building arrays},
  author    = {Carpentieri, Matteo and Robins, Alan G},
  journal   = {Journal of Wind Engineering and Industrial Aerodynamics},
  volume    = {145},
  pages     = {61--74},
  year      = {2015},
  publisher = {Elsevier}
}

@article{Tabrizi2014Performance,
  title     = {Performance and safety of rooftop wind turbines: Use of CFD to gain insight into inflow conditions},
  author    = {Tabrizi, Amir Bashirzadeh and Whale, Jonathan and Lyons, Thomas and Urmee, Tania},
  journal   = {Renewable Energy},
  volume    = {67},
  pages     = {242--251},
  year      = {2014},
  publisher = {Elsevier}
}

@article{gianfelice2022real,
  title     = {Real-time wind predictions for safe drone flights in Toronto},
  author    = {Gianfelice, Michael and Aboshosha, Haitham and Ghazal, Tarek},
  journal   = {Results in Engineering},
  volume    = {15},
  pages     = {100534},
  year      = {2022},
  publisher = {Elsevier}
}

@article{Krawczyk2025Urban,
  title     = {Urban Wind Field Effects on the Flight Dynamics of Fixed-Wing Drones},
  author    = {Krawczyk, Zack and Vuppala, Rohit KSS and Paul, Ryan and Kara, Kursat},
  journal   = {Drones},
  volume    = {9},
  number    = {5},
  pages     = {362},
  year      = {2025},
  publisher = {MDPI}
}

@article{Yazid2014A,
  title     = {A review on the flow structure and pollutant dispersion in urban street canyons for urban planning strategies},
  author    = {Yazid, Afiq Witri Muhammad and Sidik, Nor Azwadi Che and Salim, Salim Mohamed and Saqr, Khalid M},
  journal   = {Simulation},
  volume    = {90},
  number    = {8},
  pages     = {892--916},
  year      = {2014},
  publisher = {Sage Publications Sage UK: London, England}
}

@inproceedings{gnatowska2017cfd,
  title        = {CFD modelling and PIV experimental validation of flow fields in urban environments},
  author       = {Gnatowska, Renata and Sosnowski, Marcin and Uruba, V{\'a}clav},
  booktitle    = {E3S Web of Conferences},
  volume       = {14},
  pages        = {01034},
  year         = {2017},
  organization = {EDP Sciences}
}

@article{aitken2014large,
  title     = {Large eddy simulation of wind turbine wake dynamics in the stable boundary layer using the Weather Research and Forecasting Model},
  author    = {Aitken, Matthew L and Kosovi{\'c}, Branko and Mirocha, Jeffrey D and Lundquist, Julie K},
  journal   = {Journal of Renewable and Sustainable Energy},
  volume    = {6},
  number    = {3},
  year      = {2014},
  publisher = {AIP Publishing}
}

@article{Shao2023PIGNN-CFD:,
  title     = {PIGNN-CFD: A physics-informed graph neural network for rapid predicting urban wind field defined on unstructured mesh},
  author    = {Shao, Xuqiang and Liu, Zhijian and Zhang, Siqi and Zhao, Zijia and Hu, Chenxing},
  journal   = {Building and Environment},
  volume    = {232},
  pages     = {110056},
  year      = {2023},
  publisher = {Elsevier}
}

@article{Tominaga2024CFD,
  title     = {CFD simulations of turbulent flow and dispersion in built environment: A perspective review},
  author    = {Tominaga, Yoshihide},
  journal   = {Journal of Wind Engineering and Industrial Aerodynamics},
  volume    = {249},
  pages     = {105741},
  year      = {2024},
  publisher = {Elsevier}
}

@article{Hooff2017On,
  title     = {On the accuracy of CFD simulations of cross-ventilation flows for a generic isolated building: Comparison of RANS, LES and experiments},
  author    = {van Hooff, Twan and Blocken, Bert and Tominaga, Yoshihide},
  journal   = {Building and Environment},
  volume    = {114},
  pages     = {148--165},
  year      = {2017},
  publisher = {Elsevier}
}

@article{Lin2020Kriging,
  title     = {Kriging based sequence interpolation and probability distribution correction for gaussian wind field data reconstruction},
  author    = {Lin, Qiushuang and Li, Chunxiang},
  journal   = {Journal of Wind Engineering and Industrial Aerodynamics},
  volume    = {205},
  pages     = {104340},
  year      = {2020},
  publisher = {Elsevier}
}

@article{Lin2021Nonstationary,
  title     = {Nonstationary wind speed data reconstruction based on secondary correction of statistical characteristics},
  author    = {Lin, Qiushuang and Li, Chunxiang},
  journal   = {Structural Control and Health Monitoring},
  volume    = {28},
  number    = {9},
  pages     = {e2783},
  year      = {2021},
  publisher = {Wiley Online Library}
}

@article{Gao2024Urban,
  title     = {Urban wind field prediction based on sparse sensors and physics-informed graph-assisted auto-encoder},
  author    = {Gao, Huanxiang and Hu, Gang and Zhang, Dongqin and Jiang, Wenjun and Tse, KT and Kwok, KCS and Kareem, Ahsan},
  journal   = {Computer-Aided Civil and Infrastructure Engineering},
  volume    = {39},
  number    = {10},
  pages     = {1409--1430},
  year      = {2024},
  publisher = {Wiley Online Library}
}

@article{Kang2021Application,
  title     = {Application of POD reduced-order algorithm on data-driven modeling of rod bundle},
  author    = {Kang, Huilun and Tian, Zhaofei and Chen, Guangliang and Li, Lei and Wang, Tianyu},
  journal   = {Nuclear Engineering and Technology},
  volume    = {54},
  number    = {1},
  pages     = {36--48},
  year      = {2022},
  publisher = {Elsevier}
}

@article{Ti2020Wake,
  title     = {Wake modeling of wind turbines using machine learning},
  author    = {Ti, Zilong and Deng, Xiao Wei and Yang, Hongxing},
  journal   = {Applied Energy},
  volume    = {257},
  pages     = {114025},
  year      = {2020},
  publisher = {Elsevier}
}

@article{DeOliveira2022Coupling,
  title     = {Coupling a neural network technique with CFD simulations for predicting 2-D atmospheric dispersion analyzing wind and composition effects},
  author    = {de Oliveira, Jo{\~a}o Pedro Souza and Alves, Joao Victor Barbosa and Carneiro, Jo{\~a}o Neuenschwander Escosteguy and de Andrade Medronho, Ricardo and Silva, Luiz Fernando Lopes Rodrigues},
  journal   = {Journal of Loss Prevention in the Process Industries},
  volume    = {80},
  pages     = {104930},
  year      = {2022},
  publisher = {Elsevier}
}

@article{kohler2019toward,
  title     = {Toward bridging the simulated-to-real gap: Benchmarking super-resolution on real data},
  author    = {K{\"o}hler, Thomas and B{\"a}tz, Michel and Naderi, Farzad and Kaup, Andr{\'e} and Maier, Andreas and Riess, Christian},
  journal   = {IEEE transactions on pattern analysis and machine intelligence},
  volume    = {42},
  number    = {11},
  pages     = {2944--2959},
  year      = {2019},
  publisher = {IEEE}
}

@article{chowdhury2024state,
  title   = {State-of-the-Art CFD Simulation: A Review of Techniques, Validation Methods, and Application Scenarios},
  author  = {Chowdhury, Imtiaze Ahmed},
  journal = {J. Recent Trends Mech},
  volume  = {9},
  pages   = {45--53},
  year    = {2024}
}

@article{Qin2018Wind,
  title     = {Wind field reconstruction using dimension-reduction of CFD data with experimental validation},
  author    = {Qin, Li and Liu, Shi and Long, Teng and Shahzad, Muhammad Ali and Schlaberg, H Inaki and Yan, Song An},
  journal   = {Energy},
  volume    = {151},
  pages     = {272--288},
  year      = {2018},
  publisher = {Elsevier}
}

@article{Hu2024Effect,
  title     = {Effect of wind direction on natural ventilation in a multiple-room house via field measurements and numerical simulations},
  author    = {Hu, Hong and Kikumoto, Hideki and Ooka, Ryozo},
  journal   = {Journal of Wind Engineering and Industrial Aerodynamics},
  volume    = {248},
  pages     = {105718},
  year      = {2024},
  publisher = {Elsevier}
}

@article{Lawson2010Understanding,
  title     = {Understanding cavity flows using proper orthogonal decomposition and signal processing},
  author    = {Lawson, SJ and Barakos, GN and Simpson, A},
  journal   = {Journal of Algorithms \& Computational Technology},
  volume    = {4},
  number    = {1},
  pages     = {47--69},
  year      = {2010},
  publisher = {SAGE Publications Sage UK: London, England}
}

@article{Li2013Model,
  title     = {Model reduction of a coupled numerical model using proper orthogonal decomposition},
  author    = {Li, Xinya and Chen, Xiao and Hu, Bill X and Navon, I Michael},
  journal   = {Journal of Hydrology},
  volume    = {507},
  pages     = {227--240},
  year      = {2013},
  publisher = {Elsevier}
}

@article{Ferrero2018Global,
  title     = {Global and local POD models for the prediction of compressible flows with DG methods},
  author    = {Ferrero, Andrea and Iollo, Angelo and Larocca, Francesco},
  journal   = {International Journal for Numerical Methods in Engineering},
  volume    = {116},
  number    = {5},
  pages     = {332--357},
  year      = {2018},
  publisher = {Wiley Online Library}
}

@article{Miao2024Interpolation,
  title     = {Interpolation of non-stationary geo-data using Kriging with sparse representation of covariance function},
  author    = {Miao, Cong and Wang, Yu},
  journal   = {Computers and Geotechnics},
  volume    = {169},
  pages     = {106183},
  year      = {2024},
  publisher = {Elsevier}
}

@article{Risser2016Review:,
  title   = {Review: Nonstationary spatial modeling, with emphasis on process convolution and covariate-driven approaches. eprint},
  author  = {Risser, MD},
  journal = {arXiv preprint arXiv:1610.02447},
  year    = {2016}
}

@article{tang2024super,
  title     = {Super-resolution reconstruction of wind fields with a swin-transformer-based deep learning framework},
  author    = {Tang, Lingxiao and Li, Chao and Zhao, Zihan and Xiao, Yiqing and Chen, Shenpeng},
  journal   = {Physics of Fluids},
  volume    = {36},
  number    = {12},
  year      = {2024},
  publisher = {AIP Publishing}
}

@article{Li2024Wind,
  title     = {Wind Profile Reconstruction Based on Convolutional Neural Network for Incoherent Doppler Wind LiDAR},
  author    = {Li, Jiawei and Chen, Chong and Han, Yuli and Chen, Tingdi and Xue, Xianghui and Liu, Hengjia and Zhang, Shuhua and Yang, Jing and Sun, Dongsong},
  journal   = {Remote Sensing},
  volume    = {16},
  number    = {8},
  pages     = {1473},
  year      = {2024},
  publisher = {MDPI}
}

@article{tschannen2018recent,
  title   = {Recent advances in autoencoder-based representation learning},
  author  = {Tschannen, Michael and Bachem, Olivier and Lucic, Mario},
  journal = {arXiv preprint arXiv:1812.05069},
  year    = {2018}
}

@misc{vaswani2023attentionneed,
  title   = {Attention is all you need},
  author  = {Vaswani, Ashish and Shazeer, Noam and Parmar, Niki and Uszkoreit, Jakob and Jones, Llion and Gomez, Aidan N and Kaiser, {\L}ukasz and Polosukhin, Illia},
  journal = {Advances in neural information processing systems},
  volume  = {30},
  year    = {2017}
}

@article{goodfellow2014generativeadversarialnetworks,
  title     = {Generative adversarial networks},
  author    = {Goodfellow, Ian and Pouget-Abadie, Jean and Mirza, Mehdi and Xu, Bing and Warde-Farley, David and Ozair, Sherjil and Courville, Aaron and Bengio, Yoshua},
  journal   = {Communications of the ACM},
  volume    = {63},
  number    = {11},
  pages     = {139--144},
  year      = {2020},
  publisher = {ACM New York, NY, USA}
}

@article{koetzier2023deep,
  title     = {Deep learning image reconstruction for CT: technical principles and clinical prospects},
  author    = {Koetzier, Lennart R and Mastrodicasa, Domenico and Szczykutowicz, Timothy P and van der Werf, Niels R and Wang, Adam S and Sandfort, Veit and van der Molen, Aart J and Fleischmann, Dominik and Willemink, Martin J},
  journal   = {Radiology},
  volume    = {306},
  number    = {3},
  pages     = {e221257},
  year      = {2023},
  publisher = {Radiological Society of North America}
}

@article{An2024A,
  title   = {A Review of Research on Super-resolution Image Reconstruction Based on Deep Learning},
  author  = {Ning An},
  journal = {Applied and Computational Engineering},
  year    = {2024},
  volume  = {111},
  pages   = {217--224}
}

@inproceedings{Miyanawala2018A,
  title        = {A hybrid data-driven deep learning technique for fluid-structure interaction},
  author       = {Miyanawala, TP and Jaiman, Rajeev K},
  booktitle    = {International conference on offshore mechanics and arctic engineering},
  volume       = {58776},
  pages        = {V002T08A004},
  year         = {2019},
  organization = {American Society of Mechanical Engineers}
}

@article{Wang2020An,
  title     = {An end-to-end deep network for reconstructing CT images directly from sparse sinograms},
  author    = {Wang, Wei and Xia, Xiang-Gen and He, Chuanjiang and Ren, Zemin and Lu, Jian and Wang, Tianfu and Lei, Baiying},
  journal   = {IEEE Transactions on Computational Imaging},
  volume    = {6},
  pages     = {1548--1560},
  year      = {2020},
  publisher = {IEEE}
}

@article{Szczotka2019Learning,
  title     = {Learning from irregularly sampled data for endomicroscopy super-resolution: a comparative study of sparse and dense approaches},
  author    = {Szczotka, Agnieszka Barbara and Shakir, Dzhoshkun Ismail and Rav{\`\i}, Daniele and Clarkson, Matthew J and Pereira, Stephen P and Vercauteren, Tom},
  journal   = {International journal of computer assisted radiology and surgery},
  volume    = {15},
  number    = {7},
  pages     = {1167--1175},
  year      = {2020},
  publisher = {Springer}
}

@inproceedings{Makarov2018Sparse,
  title        = {Sparse depth map interpolation using deep convolutional neural networks},
  author       = {Makarov, Ilya and Korinevskaya, Alisa and Aliev, Vladimir},
  booktitle    = {2018 41st international conference on telecommunications and signal processing (TSP)},
  pages        = {1--5},
  year         = {2018},
  organization = {IEEE}
}

@article{oliver1990kriging,
  title     = {Kriging: a method of interpolation for geographical information systems},
  author    = {Oliver, Margaret A and Webster, Richard},
  journal   = {International Journal of Geographical Information System},
  volume    = {4},
  number    = {3},
  pages     = {313--332},
  year      = {1990},
  publisher = {Taylor \& Francis}
}

@inproceedings{ronneberger2015u,
  title        = {U-Net: Convolutional Networks for Biomedical Image Segmentation},
  author       = {Ronneberger, O. and Fischer, P. and Brox, T.},
  booktitle    = {International Conference on Medical image computing and computer-assisted intervention},
  pages        = {234--241},
  year         = {2015},
  organization = {Springer}
}

@article{Nowak2024Optimisation,
  title     = {Optimisation of city structures with respect to high wind speeds using U-Net models},
  author    = {Nowak, Dimitri and Werner, Jennifer and Parsons, Quentin and Johnson, Tomas and Mark, Andreas and Edelvik, Fredrik},
  journal   = {Engineering Applications of Artificial Intelligence},
  volume    = {135},
  pages     = {108812},
  year      = {2024},
  publisher = {Elsevier}
}

@inproceedings{arjovsky2017wasserstein,
  title        = {Wasserstein generative adversarial networks},
  author       = {Arjovsky, Martin and Chintala, Soumith and Bottou, L{\'e}on},
  booktitle    = {International conference on machine learning},
  pages        = {214--223},
  year         = {2017},
  organization = {PMLR}
}

@article{mirza2014conditionalgenerativeadversarialnets,
  title   = {Conditional generative adversarial nets},
  author  = {Mirza, Mehdi and Osindero, Simon},
  journal = {arXiv preprint arXiv:1411.1784},
  year    = {2014}
}

@article{dosovitskiy2020image,
  title   = {An image is worth 16x16 words: Transformers for image recognition at scale},
  author  = {Dosovitskiy, Alexey and Beyer, Lucas and Kolesnikov, Alexander and Weissenborn, Dirk and Zhai, Xiaohua and Unterthiner, Thomas and Dehghani, Mostafa and Minderer, Matthias and Heigold, Georg and Gelly, Sylvain and others},
  journal = {arXiv preprint arXiv:2010.11929},
  year    = {2020}
}

@article{Xu2021ViTAE:,
  title   = {Vitae: Vision transformer advanced by exploring intrinsic inductive bias},
  author  = {Xu, Yufei and Zhang, Qiming and Zhang, Jing and Tao, Dacheng},
  journal = {Advances in neural information processing systems},
  volume  = {34},
  pages   = {28522--28535},
  year    = {2021}
}

@article{cheng2025machine,
  title     = {Machine learning for modelling unstructured grid data in computational physics: a review},
  author    = {Cheng, Sibo and Bocquet, Marc and Ding, Weiping and Finn, Tobias Sebastian and Fu, Rui and Fu, Jinlong and Guo, Yike and Johnson, Eleda and Li, Siyi and Liu, Che and others},
  journal   = {Information Fusion},
  pages     = {103255},
  year      = {2025},
  publisher = {Elsevier}
}

@article{cheng2024efficient,
  title     = {Efficient deep data assimilation with sparse observations and time-varying sensors},
  author    = {Cheng, Sibo and Liu, Che and Guo, Yike and Arcucci, Rossella},
  journal   = {Journal of Computational Physics},
  volume    = {496},
  pages     = {112581},
  year      = {2024},
  publisher = {Elsevier}
}

@article{gao2024uncertainties,
  title     = {Uncertainties in temperature statistics and fluxes determined by sonic anemometers due to wind-induced vibrations of mounting arms},
  author    = {Gao, Zhongming and Liu, Heping and Li, Dan and Yang, Bai and Walden, Von and Li, Lei and Bogoev, Ivan},
  journal   = {Atmospheric Measurement Techniques},
  volume    = {17},
  number    = {13},
  pages     = {4109--4120},
  year      = {2024},
  publisher = {Copernicus Publications G{\"o}ttingen, Germany}
}

@article{riva2024multi,
  title     = {Multi-physics model bias correction with data-driven reduced order techniques: Application to nuclear case studies},
  author    = {Riva, Stefano and Introini, Carolina and Cammi, Antonio},
  journal   = {Applied Mathematical Modelling},
  volume    = {135},
  pages     = {243--268},
  year      = {2024},
  publisher = {Elsevier}
}

@article{cammi2024data,
  title     = {Data-driven model order reduction for sensor positioning and indirect reconstruction with noisy data: Application to a Circulating Fuel Reactor},
  author    = {Cammi, Antonio and Riva, Stefano and Introini, Carolina and Loi, Lorenzo and Padovani, Enrico},
  journal   = {Nuclear Engineering and Design},
  volume    = {421},
  pages     = {113105},
  year      = {2024},
  publisher = {Elsevier}
}

@article{manohar2018data,
  title     = {Data-driven sparse sensor placement for reconstruction: Demonstrating the benefits of exploiting known patterns},
  author    = {Manohar, Krithika and Brunton, Bingni W and Kutz, J Nathan and Brunton, Steven L},
  journal   = {IEEE Control Systems Magazine},
  volume    = {38},
  number    = {3},
  pages     = {63--86},
  year      = {2018},
  publisher = {IEEE}
}

@article{wang2024dynamical,
  title     = {Dynamical system prediction from sparse observations using deep neural networks with Voronoi tessellation and physics constraint},
  author    = {Wang, Hanyang and Zhou, Hao and Cheng, Sibo},
  journal   = {Computer Methods in Applied Mechanics and Engineering},
  volume    = {432},
  pages     = {117339},
  year      = {2024},
  publisher = {Elsevier}
}

@article{fukami2021global,
  title     = {Global field reconstruction from sparse sensors with Voronoi tessellation-assisted deep learning},
  author    = {Fukami, Kai and Maulik, Romit and Ramachandra, Nesar and Fukagata, Koji and Taira, Kunihiko},
  journal   = {Nature Machine Intelligence},
  volume    = {3},
  number    = {11},
  pages     = {945--951},
  year      = {2021},
  publisher = {Nature Publishing Group UK London}
}

@article{gao2024sigan,
  title     = {SiGAN: A 3D sensor importance deep generative model for urban wind flow field monitoring},
  author    = {Gao, Huanxiang and Hu, Gang and Zhang, Dongqin and Jiang, Wenjun and Tse, KT and Noack, Bernd R},
  journal   = {Building and Environment},
  volume    = {262},
  pages     = {111787},
  year      = {2024},
  publisher = {Elsevier}
}

@article{gao2023optimal,
  title     = {An optimal sensor placement scheme for wind flow and pressure field monitoring},
  author    = {Gao, Huanxiang and Liu, Junle and Lin, Pengfei and Hu, Gang and Patruno, Luca and Xiao, Yiqing and Tse, KT and Kwok, KCS},
  journal   = {Building and Environment},
  volume    = {244},
  pages     = {110803},
  year      = {2023},
  publisher = {Elsevier}
}

@article{hou2024machine,
  title     = {Machine-learned flow estimation with sparse data—Exemplified for the rooftop of an unmanned aerial vehicle vertiport},
  author    = {Hou, Chang and Marra, Luigi and Cornejo Maceda, Guy Y and Jiang, Peng and Chen, Jingguo and Liu, Yutong and Hu, Gang and Chen, Jialong and Ianiro, Andrea and Discetti, Stefano and others},
  journal   = {Physics of Fluids},
  volume    = {36},
  number    = {12},
  year      = {2024},
  publisher = {AIP Publishing}
}

@article{fan2025vitae,
  title     = {ViTAE-SL: A vision transformer-based autoencoder and spatial interpolation learner for field reconstruction},
  author    = {Fan, Hongwei and Cheng, Sibo and de Nazelle, Audrey J and Arcucci, Rossella},
  journal   = {Computer Physics Communications},
  volume    = {308},
  pages     = {109464},
  year      = {2025},
  publisher = {Elsevier}
}

@article{liu2024application,
  title={Application and comparison of several adaptive sampling algorithms in reduced order modeling},
  author={Liu, Xirui and Wang, Zhiyong and Ji, Hongjun and Gong, Helin},
  journal={Heliyon},
  volume={10},
  number={15},
  year={2024},
  publisher={Elsevier}
}

@article{xiao2019reduced,
  title={A reduced order model for turbulent flows in the urban environment using machine learning},
  author={Xiao, Dunhui and Heaney, CE and Mottet, Laetitia and Fang, F and Lin, W and Navon, IM and Guo, Y and Matar, OK and Robins, AG and Pain, CC},
  journal={Building and Environment},
  volume={148},
  pages={323--337},
  year={2019},
  publisher={Elsevier}
}

@article{eidi2022data,
  title={Data-driven quantification of model-form uncertainty in Reynolds-averaged simulations of wind farms},
  author={Eidi, Ali and Zehtabiyan-Rezaie, Navid and Ghiassi, Reza and Yang, Xiang and Abkar, Mahdi},
  journal={Physics of Fluids},
  volume={34},
  number={8},
  year={2022},
  publisher={AIP Publishing}
}

@article{formont2025evaluation,
  title={Evaluation of data-driven models for post-combustion CO2 capture: A comparative analysis of accuracy, robustness and feasibility},
  author={Formont, Valentin and Rasheed, Adil and Moser, Peter and Wiechers, Georg and Nord, Lars O},
  journal={International Journal of Greenhouse Gas Control},
  volume={146},
  pages={104450},
  year={2025},
  publisher={Elsevier}
}

@article{kaiser2024cluster,
  title={Cluster-based Bayesian approach for noisy and sparse data: application to flow-state estimation},
  author={Kaiser, Frieder and Iacobello, Giovanni and Rival, David E},
  journal={Proceedings of the Royal Society A: Mathematical, Physical and Engineering Sciences},
  volume={480},
  number={2291},
  year={2024},
  publisher={The Royal Society}
}
\end{document}